\documentclass{article}
\usepackage[T1]{fontenc}
\usepackage{MyProject}
\usepackage[T1]{fontenc}
\ifdefined\XeTeXversion\else
\usepackage[utf8]{inputenc}
\fi
\usepackage{amsmath,amssymb,amsfonts,bm}
\usepackage{booktabs,wrapfig,float,placeins,needspace}
\usepackage{makecell,multirow,tabularx,array}
\usepackage{fvextra,listings,caption,pifont}

\usepackage{amsmath,amsfonts,bm}

\def\eqref#1{equation~\ref{#1}}

\def\1{\bm{1}}

\DeclareMathAlphabet{\mathsfit}{\encodingdefault}{\sfdefault}{m}{sl}
\SetMathAlphabet{\mathsfit}{bold}{\encodingdefault}{\sfdefault}{bx}{n}

\newtcblisting{promptbox}[1]{
    breakable,
    colback=gray!5,
    colframe=gray!40,
    boxrule=0.5pt,
    arc=1pt,
    left=4pt,
    right=4pt,
    top=4pt,
    bottom=4pt,
    title=\textbf{#1},
    fonttitle=\bfseries,
    coltitle=black,
    listing only,
    listing options={
        basicstyle=\ttfamily\scriptsize,
        breaklines=true,
        breakatwhitespace=true,
        columns=fullflexible,
        keepspaces=true,
        showstringspaces=false,
        upquote=true,
        escapeinside={(*@}{@*)}
    }
}
\definecolor{methodblue}{RGB}{255,243,230}
\newcommand{\method}{CheckDistill}

\ReportShowAuthorstrue
\title{CheckDistill: Adaptive Checklist Distillation for Efficient Distributional Reward Modeling in Image Editing}
\author{%
  Yaqi Cai\textsuperscript{*}\quad
  Zuan Gao\textsuperscript{*}\quad
  Mengting Chen\textsuperscript{\textdagger}\quad
  Wei Li\quad
  Bo Zheng\textsuperscript{\EnvelopeSymbol}%
}
\renewcommand{\ReportEqualContributionNote}{Equal Contribution.}
\renewcommand{\ReportInternshipNote}{Work done during Yaqi Cai's internship at Alibaba.}
\renewcommand{\ReportProjectLeadNote}{Project Leader.}
\renewcommand{\ReportCorrespondingNote}{Corresponding Author.}
\renewcommand{\ReportContactEmail}{cyaqi@mail.ustc.edu.cn}
\renewcommand{\ReportAffiliations}{Taobao \& Tmall group of Alibaba}
\renewcommand{\ProjectPageURL}{https://faltingsa.github.io/CheckDistill.github.io/}
\hypersetup{
  pdftitle={CheckDistill: Adaptive Checklist Distillation for Efficient Distributional Reward Modeling in Image Editing},
  pdfsubject={CheckDistill Technical Report}
}
\ifReportShowAuthors
  \hypersetup{pdfauthor={Yaqi Cai, Zuan Gao, Mengting Chen, Wei Li, Bo Zheng}}
\else
  \hypersetup{pdfauthor={}}
\fi

\begin{document}
\maketitle
\begin{abstract}
Reinforcement learning for instruction-guided image editing requires reward models that provide reliable, accurate, and efficient feedback. Direct scalar prediction compresses multidimensional editing preferences, whereas generative evaluators can expose their judgment through explicit reasoning. However, broad scoring criteria do not specify all instance-dependent editing and preservation constraints, and generating long reasoning trajectories is costly for online optimization. We introduce \method, a pointwise reward modeling framework that internalizes instruction-adaptive checklists. We decompose each editing condition into atomic, independently verifiable requirements and construct four-dimensional scoring labels together with structured verification evidence. After supervised score initialization, a privileged teacher conditions on checklists, item-level judgments, and dimension-level rationales to supervise the student's own scoring trajectories. A reliability-weighted on-policy self-distillation objective adjusts supervision at score tokens according to agreement between the teacher's expected scores and reference labels, while retaining candidate probability weights in a truncated token support. The deployed student directly predicts four score distributions without checklist inputs or explicit reasoning and aggregates their expectations into a scalar reward. \method\ achieves aggregate accuracies of 72.30\%, 56.67\% and 82.90\% on MMRB2, MER-Bench and EditReward-Bench, respectively, exceeding the compared open-source baselines on all three metrics. 
\end{abstract}

\section{Introduction}
\label{sec:introduction}

Instruction-guided image editing is a key application of multi-modal generation, enabling users to perform diverse image modifications through natural language instructions. Reinforcement learning can substantially improve image editing performance, with reward models playing a critical role in guiding optimization. An effective reward model must evaluate candidate results across complementary dimensions, distinguish subtle yet consequential differences, and provide accurate preference rankings and score singals. Frequent reward queries during online reinforcement learning further require reward models to combine comprehensive, fine-grained evaluation with efficient scoring.

\begin{figure}[!htbp]
\centering
\includegraphics[width=\linewidth]{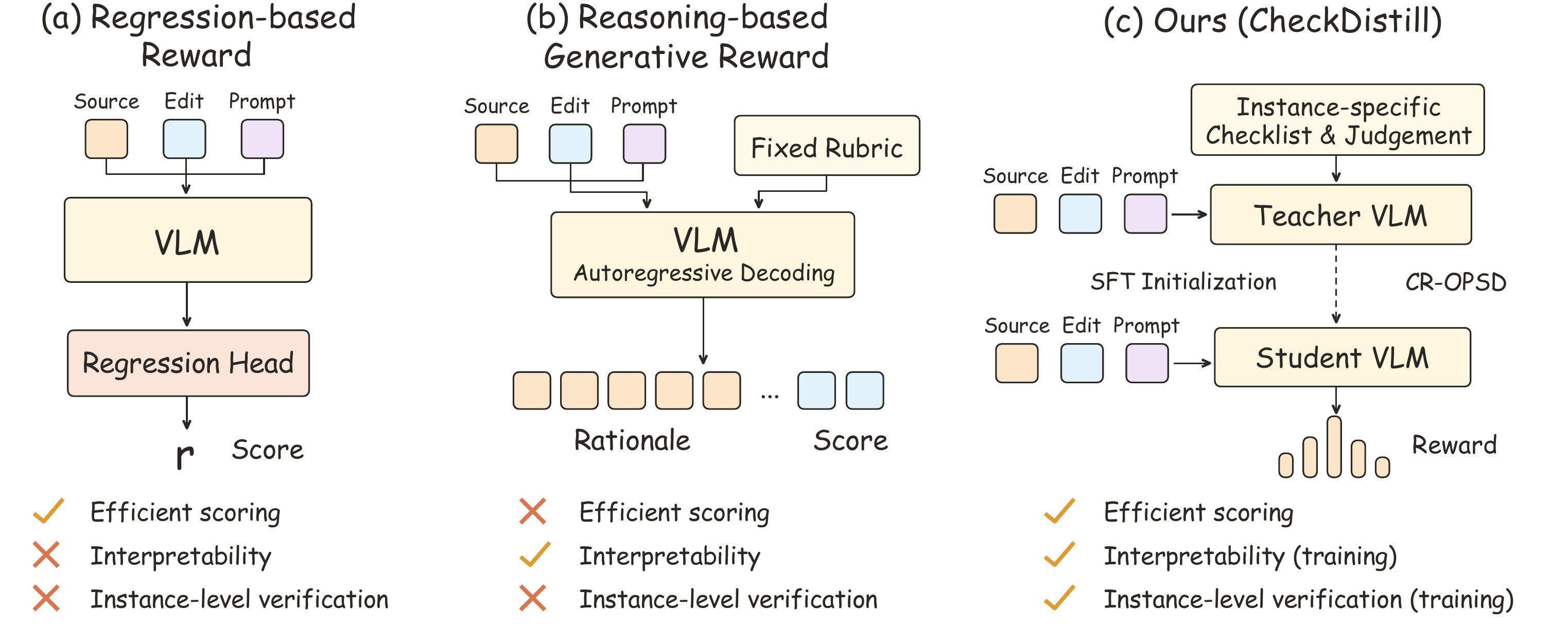}
\caption{\textbf{Overview} of CheckDistill compared with existing reward model architecture.}
\label{fig:intro}
\end{figure}

Existing reward modeling approaches do not fully meet these requirements. As shown in Fig.~\ref{fig:intro}(a), regression-based reward models~\citep{xu2023imagereward,wu2023hpsv2,wu2026editreward} enable efficient scoring. However, these models rely on coarse-grained subjective preferences, typically learned through pairwise objectives such as the Bradley--Terry loss~\citep{bradley1952rank}. These objectives alone provide neither explicit verification evidence nor absolute score calibration. These limitations reduce interpretability and may make score distributions more prone to drift and the models more vulnerable to reward hacking. Fig.~\ref{fig:intro}(b) shows the reasoning-based generative reward models~\citep{luo2026editscore,wang2026rationalrewards,kan2026rubricrm}. This paradigm uses fixed rubric prompts to guide evaluation through explicit reasoning before scoring. However, even when rubrics are generated adaptively, dimension-level prompts may leave concrete verification targets insufficiently specified. The resulting reasoning may not systematically cover the instance-specific editing and preservation constraints that distinguish otherwise similar candidates. Moreover, reasoning-based scoring requires additional autoregressive decoding at each evaluation. Although downstream optimization uses the resulting reward scores, generating the accompanying rationales adds latency and limits throughput during online RL training.

These limitations motivate two complementary principles. First, \textbf{Dynamic Checklist Scaling}: reward judgments should be grounded in instance-adaptive identification of editing and preservation requirements, decomposed into fine-grained, atomic, and independently verifiable checks. Expanding the coverage and specificity of these checks can support more accurate and robust reward judgments. Second, \textbf{Context-Compressed Reward Scoring}: efficient online deployment requires reducing the decoding overhead of lengthy chain-of-thought reasoning context while preserving reward quality. Therefore, the core challenge is to align score distributions under standard editing inputs with their evidence-conditioned counterparts, while reducing inference-time dependence on extensive verification contexts and explicit reasoning.

Guided by these principles, we propose \textbf{CheckDistill}, an image editing reward modeling framework that combines dynamic checklist-based data construction with reliability-aware on-policy self-distillation, as illustrated in Fig.~\ref{fig:intro}(c). The framework builds structured verification evidence during training and transfers evidence-enhanced judgments into efficient direct scoring.

\begin{figure}[!t]
\centering
\includegraphics[width=\linewidth]{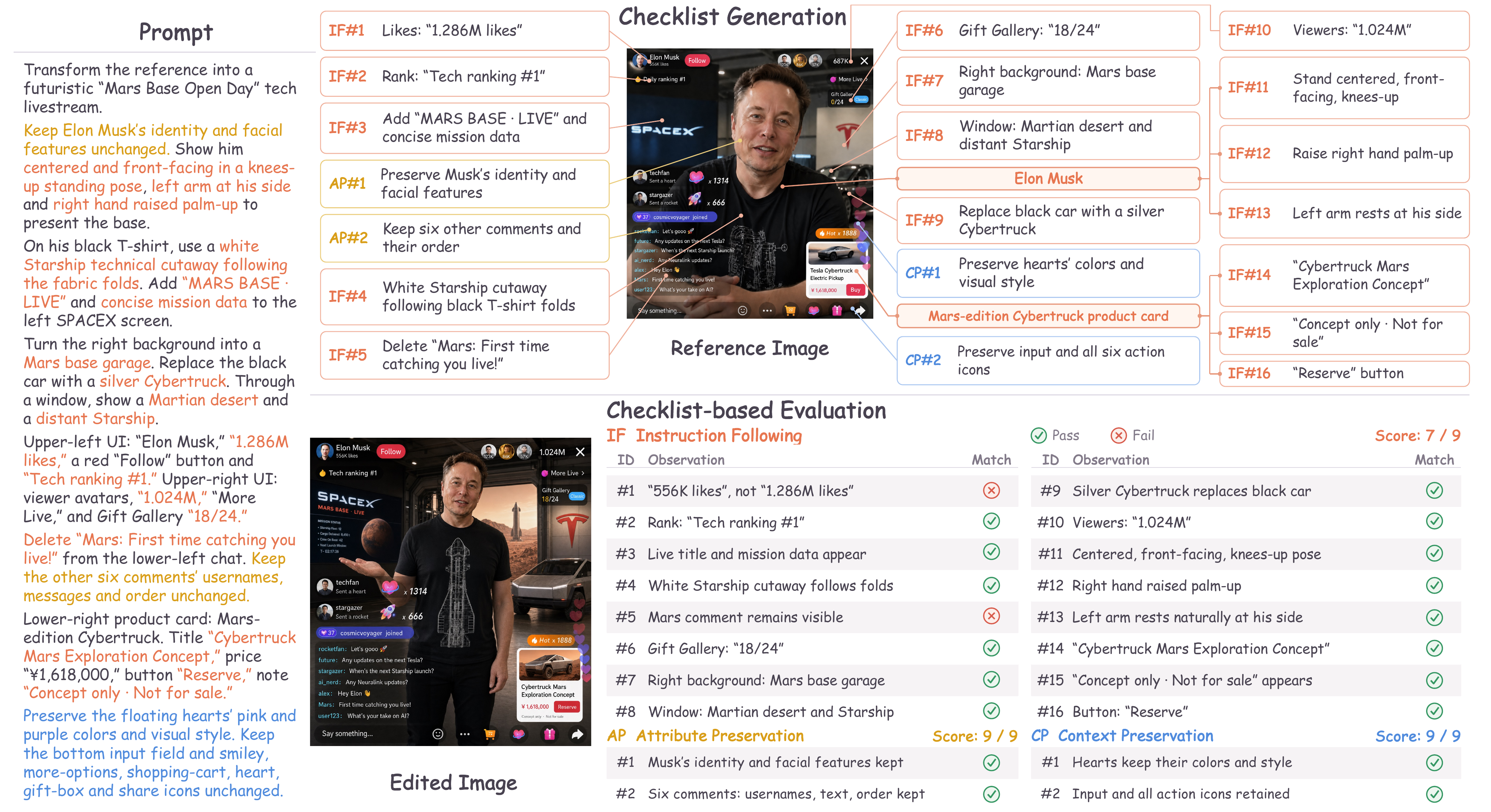}
\caption{An example of checklist generation and checklist-based evaluation for image editing.}
\label{fig:intro-case}
\end{figure}

For data construction, Fig.~\ref{fig:intro-case} illustrates how the source image and editing instruction guide atomic checks for Instruction Following (IF), Attribute Preservation (AP), and Context Preservation (CP). These checks are shared across candidates under the same editing condition. In the illustrated example, item-level judgments identify an unchanged like count and an undeleted comment, while separately verifying preservation requirements. Perceptual Quality (PQ) is assessed independently using a fixed general rubric. Combining EditReward-Data~\citep{wu2026editreward} with additional Qwen-Image-Edit~\citep{wu2025qwenimage} candidates, we construct \textbf{CheckDistill-40K} through this annotation pipeline and human preference-consistency filtering, pairing four-dimensional score labels with item-level judgments and dimension-level rationales.

CheckDistill adopts two-stage training. Mixed SFT first establishes direct scoring and checklist-guided reasoning capabilities. Then we introduce \textbf{Checklist-Guided Reliability-Aware On-Policy Self-Distillation (CR-OPSD)}.
The teacher additionally receives checklists, item-level judgments, and dimension-level rationales, providing distributional supervision on response prefixes generated by a student that observes only standard editing inputs. Since evidence-enhanced predictions may still be unreliable, we use agreement between the teacher's expected score and the reference label as a reliability proxy to modulate distillation at the four score positions. At inference time, only the student is retained, and expected scores from its four-dimensional score distributions are aggregated into a scalar reward, without additional verification context or lengthy rationale generation.

CheckDistill achieves aggregate preference accuracies of
72.30\% on MMRB2,
56.67\% on MER-Bench, and
82.90\% on EditReward-Bench, outperforming the
compared open-source baselines across all three benchmarks. 
When used as the reward model for DiffusionNFT~\citep{zheng2025diffusionnft} post-training of Qwen-Image-Edit, it achieves scores of 4.46 on ImgEdit-Bench, 4.26 on REDEdit-Bench, and 8.28 on GEdit-EN.

Our main contributions are as follows:
\begin{itemize}
    \item \textbf{Dynamic checklist-based reward supervision.}
    We construct \textbf{CheckDistill-40K}, pairing four-dimensional scores with instance-specific verification evidence through dynamic checklists and human preference-consistency filtering.

    \item \textbf{Reliability-aware distillation for efficient reward scoring.}
    We propose \textbf{CR-OPSD} to transfer evidence-conditioned scoring behavior into a direct reward model through reliability-aware on-policy distribution alignment.

    \item \textbf{Leading reward accuracy and downstream editing gains.}
    CheckDistill outperforms the compared state-of-the-art open-source reward models on image editing reward benchmarks and improves Qwen-Image-Edit through reward-guided post-training.
\end{itemize}

\KeyInsight{Scaling instance-specific dynamic checklists defines the frontier of reward quality; compressing reasoning enables deployment at scale.}

\section{Related Work}
Reinforcement learning (RL)~\citep{black2024ddpo,fan2023dpok,liu2025flowgrpo,xue2025dancegrpo,zheng2025diffusionnft} has emerged as an important post-training paradigm for visual generation, enabling diffusion and flow-based models~\citep{rombach2022ldm,podell2023sdxl,esser2024sd3,blackforestlabs2025flux} to achieve stronger alignment. Its effectiveness, however, relies heavily on accurate and efficient reward models.

\paragraph{Regression-based Reward Modeling.}
Regression-based visual reward models learn human preferences from vision-language representations using regression or Bradley--Terry-style preference objectives~\citep{wu2023hps,xu2023imagereward,kirstain2023pickscore,wu2023hpsv2,zhang2024mps,wu2024visionprefer,zhang2024hive,chen2025adiee,ma2025hpsv3,wang2025unifiedreward,wu2026editreward}. These methods are efficient at inference, but typically output scalar scores without explicit reasoning, limiting interpretability and diagnostic feedback for fine-grained evaluation.

\paragraph{Generative Reward Modeling.}
Generative reward models such as EditScore~\citep{luo2026editscore}, RationalRewards~\citep{wang2026rationalrewards}, SpatialReward~\citep{long2026spatialreward}, and UnifiedReward-Think~\citep{wang2025unifiedrewardthink} produce explicit rationales or chain-of-thought reasoning for fine-grained reward prediction under fixed evaluation criteria. However, these criteria are not instantiated into concrete, input-specific verification requirements, and inference-time reasoning introduces substantial overhead in large-scale online RL. JRM~\citep{yang2026jrm} and Z-Reward~\citep{jin2026zreward} improve efficiency by internalizing reasoning into direct reward prediction, but still lack instance-adaptive verification criteria.

Instance-adaptive methods tailor evaluation criteria to each input. RubricRM~\citep{kan2026rubricrm} generates input-specific dimensions and scoring criteria, but remains focused on high-level evaluation dimensions. Edit-R1~\citep{guo2026editr1} decomposes editing instructions into sample-specific principles for finer-grained verification, but still requires explicit reasoning at inference time. Checklist-based methods translate evaluation requirements into directly verifiable items~\citep{cook2024tick}; rDPO~\citep{yu2026rdpo}, for example, constructs instance-specific checklist-style rubrics primarily for preference data construction. In contrast, CheckDistill uses source-image-grounded atomic checklists only as privileged information during reward-model training, distilling fine-grained instance-specific verification requirements into direct reward prediction without checklist generation or explicit reasoning at inference time.

\section{Method}
\label{sec:method}
\begin{figure}[!t]
\centering
\includegraphics[width=\linewidth]{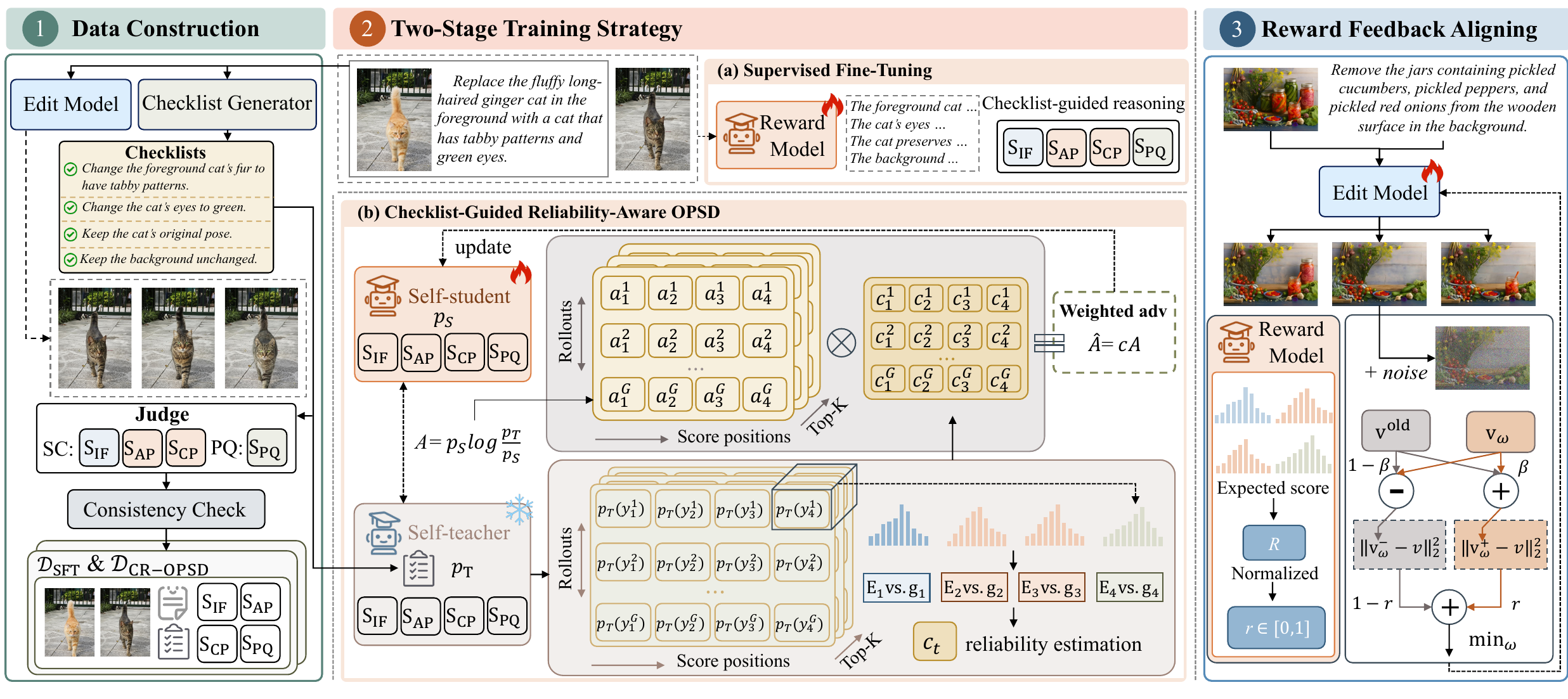}
\caption{\textbf{Overview of CheckDistill}, comprising reward data construction, two-stage reward model training via supervised fine-tuning and checklist-guided reliability-aware OPSD, and reward feedback alignment for image editing.}
\label{fig:method}
\end{figure}

As shown in Figure~\ref{fig:method}, we propose \method, a generative reward model that leverages instance-specific checklists for fine-grained evaluation and internalizes checklist-derived evaluation knowledge through CR-OPSD. Given an input $x=(u, I^{\mathrm{src}}, I^{\mathrm{edt}})$, CheckDistill maps it to a multi-dimensional scoring response $y$, where $u$ denotes the editing instruction, $I^{\mathrm{src}}$ the source image, and $I^{\mathrm{edt}}$ the edited image.

\subsection{Evaluation Formulation}

\paragraph{Evaluation Dimensions.}
We evaluate image editing from two complementary aspects: \emph{Semantic Consistency} (SC) and \emph{Perceptual Quality} (PQ). SC comprises three dimensions: \emph{Instruction Following} (IF), which measures whether the requested edits are correctly executed; \emph{Attribute Preservation} (AP), which measures whether unspecified attributes of edited objects are preserved; and \emph{Context Preservation} (CP), which measures whether non-edited content remains unchanged. 
PQ measures the overall visual quality of the edited image. 
All four scores, $S_{\mathrm{IF}}$, $S_{\mathrm{AP}}$, $S_{\mathrm{CP}}$, and $S_{\mathrm{PQ}}$, are defined on a discrete scale from $0$ to $9$.

\paragraph{Multidimensional Reward Aggregation.}
To obtain a scalar feedback signal for downstream post-training of the editing model, we account for the varying importance of different dimensions by first aggregating the three SC scores using dimension-specific weights, and then combining SC and PQ through a weighted geometric mean:
\begin{equation}
R_{\mathrm{SC}}(x)
=
\sum_{\delta\in\mathcal{D}_{\mathrm{SC}}}
w_\delta S_\delta(x),
\qquad
R(x)
=
R_{\mathrm{SC}}(x)^\lambda
S_{\mathrm{PQ}}(x)^{1-\lambda}.
\label{eq:final_reward}
\end{equation}
where $\sum_{\delta\in\mathcal{D}_{\mathrm{SC}}} w_\delta=1$ and $w_\delta \geq0$, with $\mathcal{D}_{\mathrm{SC}}=\{\mathrm{IF},\mathrm{AP},\mathrm{CP}\}$, and $\lambda\in[0,1]$ controls the relative contribution of semantic consistency and perceptual quality.

\subsection{Reward Data Construction Pipeline}
\label{sec:data_construction}

To construct the reward training data, we develop an annotation pipeline that leverages instance-specific checklists to facilitate fine-grained evaluation and reliable score annotation, while also providing structured privileged information for subsequent knowledge internalization via CR-OPSD.

\paragraph{Candidate Sampling and Human Preferences.}
\label{sec:candidate_expansion}
Given an editing condition $h=(u,I^{\mathrm{src}})$, we generate multiple edited images with an image editing model using different random seeds, forming a diverse candidate pool $\mathcal{I}^{\mathrm{edt}}=\left\{I_k^{\mathrm{edt}}\right\}_{k=1}^{K}$.
Human annotators then perform pairwise comparisons among the candidates. For each annotated pair, the preferred and non-preferred candidates are denoted as $I_w^{\mathrm{edt}}$ and $I_l^{\mathrm{edt}}$, respectively, yielding the human preference relation $I_w^{\mathrm{edt}} \succ I_l^{\mathrm{edt}}$.

\paragraph{Hybrid Rubric Design.}
Our hybrid rubric consists of an instance-specific checklist-based rubric for SC and a fixed general rubric for PQ, with the detailed rubric templates provided in Appendix~\ref{app:en:annotation}.

\label{sec:rubric_construction}
\noindent\textbf{(1) Instance-specific checklist for SC.}
To provide more fine-grained and traceable evidence than direct dimension-level assessment, we decompose the three SC dimensions into an instance-specific checklist of atomic verification items, each grounded in concrete objects, attributes, or regions.
Specifically, for each editing condition $h$, we use Gemini-3.0-Flash to generate a variable-length checklist
$C=\{c_j\}_{j=1}^{M}$,
without access to any candidate result.
where $M$ denotes the number of checklist items. 
The same checklist $C$ is shared across all candidates under the same editing condition, ensuring that they are evaluated against a consistent set of criteria.

\noindent\textbf{(2) General rubric for PQ.}
Since PQ concerns generic visual quality rather than instance-specific editing constraints, we adopt a fixed rubric shared across all samples and assess PQ directly from the edited image without reference to the source image. The rubric focuses on key perceptual factors such as realism and overall naturalness.

\paragraph{Independent SC and PQ Judgment.}
\label{sec:scoring}
We employ two independent Gemini-3.0-Flash annotation templates (see Appendix~\ref{app:en:annotation}) for SC and PQ assessment.
For the $k$-th candidate, the SC annotator takes $(x_k, C)$ as input and verifies each checklist item according to its associated dimension, producing item-level judgments $J_k$. These judgments are then summarized into dimension-level rationales $A_k^{\mathrm{SC}}$ and scores $\mathbf{g}_k^{\mathrm{SC}}=(g_k^{\mathrm{IF}},g_k^{\mathrm{AP}},g_k^{\mathrm{CP}})$.
The PQ annotator assesses the visual quality of the edited result from $(u, I_k^{\mathrm{edt}})$ and outputs a rationale $A_k^{\mathrm{PQ}}$ and score $g_k^{\mathrm{PQ}}$.

\paragraph{Preference-Consistency Filtering.}
\label{sec:preference_filtering}
We use the human preferences collected in Sec.~\ref{sec:candidate_expansion} as an external reference to verify the ranking consistency of the automatically annotated labels. We perform filtering at the candidate-group level. For each annotated pair with $I_w^{\mathrm{edt}} \succ I_l^{\mathrm{edt}}$, we aggregate the four-dimensional scores according to Eq.~\ref{eq:final_reward} and regard the pair as preference-consistent when $R_w > R_l$. Since each candidate may participate in multiple pairwise comparisons, we retain the largest subset whose induced ranking agrees with all available human preferences among the retained candidates.

Through the above pipeline, we construct the \textbf{CheckDistill-40K} dataset, comprising approximately 20K samples from EditReward-Data~\citep{wu2026editreward} and 20K samples generated with Qwen-Image-Edit following Sec.~\ref{sec:candidate_expansion}. 
For the first-stage SFT, we use a 10K subset, denoted as $\mathcal{D}_{\mathrm{SFT}}$, and mix two supervision formats: (1) given $x_k$ as input, the model predicts the four-dimensional scores $\mathbf{g}_k=[g_k^{\mathrm{IF}},g_k^{\mathrm{AP}},g_k^{\mathrm{CP}},g_k^{\mathrm{PQ}}]$; and (2) given $(x_k,C)$ as input, the model is supervised to generate the checklist judgments $J_k$, rationales $A_k=(A_k^{\mathrm{SC}},A_k^{\mathrm{PQ}})$, and scores $\mathbf{g}_k$. For the second-stage CR-OPSD training, we use the full dataset
$\mathcal{D}_{\mathrm{CR\text{-}OPSD}}=\{(x_k,\mathbf{g}_k,z_k)\}_{k=1}^{N}$,
where $z_k=(C,J_k,A_k)$ denotes the structured privileged information.

\subsection{CheckDistill Training and Downstream Alignment}
\label{sec:training}

\paragraph{Stage I: Supervised Initialization.}
\label{sec:sft}
We first perform supervised fine-tuning to initialize CheckDistill with checklist-guided reasoning and the ability to directly predict
four-dimensional scores from the editing instruction, source image, and edited result. Let $p_\theta(y\mid x')$ denote the conditional
autoregressive distribution over responses defined by CheckDistill with parameters $\theta$, where $x'$ denotes the input for the corresponding training format. The training objective is
\begin{equation}
\mathcal{L}_{\mathrm{SFT}}(\theta)
=
-\mathbb{E}_{(x',y)\sim\mathcal{D}_{\mathrm{SFT}}}
\left[
\sum_{t=1}^{|y|}
\log p_\theta(y_t\mid x',y_{<t})
\right].
\label{eq:sft_loss}
\end{equation}
where $(x',y)\in\{(x,\mathbf{g}),\,((x,C),(J,A,\mathbf{g}))\}$.
This stage elicits checklist-guided reasoning and establishes the score range and output structure, providing a foundation for CR-OPSD.

\paragraph{Stage II: Checklist-Guided Reliability-Aware OPSD (CR-OPSD).}
\label{sec:self_distillation}

We then propose CR-OPSD to internalize fine-grained checklist-guided reasoning into the reward model for efficient inference. Specifically, CR-OPSD builds on OPSD~\citep{zhao2026opsd} by using the structured evidence $z=(C,J,A)$ as privileged information and introducing reliability-aware weighting to reduce the influence of inaccurate teacher predictions. 
Both the student and teacher are initialized from the SFT checkpoint.
Their predictive distributions are defined as
\begin{equation}
p_S(y_t\mid x,y_{<t})
=
p_{\theta}(y_t\mid x,y_{<t}),
\qquad
p_T(y_t\mid x,z,y_{<t})
=
p_{\theta}(y_t\mid x,z,y_{<t}).
\end{equation}

Let $\mathcal{T}_t=\operatorname{TopK}(p_S(\cdot\mid x,y_{<t}))$ denote the student's Top-$K$ tokens at position $t$. We define $\mathcal{S}=\{0,\ldots,9\}$ as the set of valid score tokens and $\mathcal{S}_t=\mathcal{S}\cap\mathcal{T}_t$. At each score position, we adopt a Q-Align-style~\citep{wu2023qalign} probabilistic readout over $\mathcal{S}_t$ to compute the teacher's expected score:
\begin{equation}
\bar{p}_T(s\mid x,z,y_{<t})
=
\frac{p_T(s\mid x,z,y_{<t})}
{\sum_{s'\in\mathcal{S}_t}p_T(s'\mid x,z,y_{<t})},
\qquad
E_t
=
\sum_{s\in\mathcal{S}_t}
s\,\bar{p}_T(s\mid x,z,y_{<t}).
\label{eq:teacher_expected_score}
\end{equation}

Let $\delta(t)$ denote the scoring dimension associated with position $t$, with annotated score $g^{\delta(t)}$. We measure teacher--label agreement as
\begin{equation}
r_{\mathrm{unit},t}
=
\operatorname{clip}\!\left(
1-\frac{|E_t-g^{\delta(t)}|}{9},
0,1
\right),
\qquad
r_t
=
r_{\min}
+
(r_{\max}-r_{\min})
\frac{1-\cos(\pi r_{\mathrm{unit},t})}{2}.
\label{eq:score_reliability}
\end{equation}
The position-wise coefficient is defined as $c_t=(1-m_t)+m_t r_t$, where $m_t=1$ at the four score positions and $m_t=0$ otherwise.
For each $y_t\in\mathcal{T}_t$, the reliability-weighted advantage is
\begin{equation}
\hat{A}_t(y_t)
=
\operatorname{sg}\!\left[
c_t\,
p_S(y_t\mid x,y_{<t})
\log
\frac{
p_T(y_t\mid x,z,y_{<t})
}{
p_S(y_t\mid x,y_{<t})
}
\right],
\qquad
y_t\in\mathcal{T}_t.
\label{eq:weighted_advantage}
\end{equation}
where $\operatorname{sg}[\cdot]$ denotes the stop-gradient operator.
Aggregating over the Top-$K$ candidates at each position gives the final objective:
\begin{equation}
\mathcal{L}_{\mathrm{CR-OPSD}}(\theta)
=
-\mathbb{E}\!\left[
\frac{1}{|y|}
\sum_t
\sum_{y_t\in\mathcal{T}_t}
\hat{A}_t(y_t)
\log p_S(y_t\mid x,y_{<t})
\right].
\label{eq:checkdistill_loss}
\end{equation}

\paragraph{Reward Feedback for Alignment.}
\label{sec:reward_feedback}
We use CheckDistill to guide RL post-training of the image editing model with DiffusionNFT~\citep{zheng2025diffusionnft}. For policy optimization, the reward of each candidate is computed from CheckDistill's predicted score distributions. Specifically, rather than directly taking the most likely discrete score token, we derive a continuous score for each dimension from the predicted distribution. For $\delta\in\mathcal{D}_{\mathrm{SC}}\cup\{\mathrm{PQ}\}$, let $t_\delta$ denote its corresponding score position. The reward model probabilities over valid score tokens are renormalized to compute the expected score:
\begin{equation}
\bar{p}_S(s\mid x,y_{<t_\delta})
=
\frac{
p_S(s\mid x,y_{<t_\delta})
}{
\sum_{s'\in\mathcal{S}}
p_S(s'\mid x,y_{<t_\delta})
},
\qquad
\bar{S}_\delta(x)
=
\sum_{s\in\mathcal{S}}
s\,\bar{p}_S(s\mid x,y_{<t_\delta}),
\quad s\in\mathcal{S}.
\label{eq:expected_score}
\end{equation}
This distribution-based readout preserves uncertainty over the ordered score space, enabling finer-grained reward differentiation even among candidates with the same most likely score token. The resulting expected dimensional scores are then aggregated into the final reward using Eq.~\ref{eq:final_reward}, which serves as the reward signal for DiffusionNFT-based policy optimization.

\section{Experiments}
\label{sec:experiments}
\subsection{Performance on Reward Modeling Benchmarks}

\paragraph{Implementation Details.}
We implement \method\ on Qwen3-VL-8B-Instruct~\citep{bai2025qwen3vl}, Qwen3.5-4B, and Qwen3.5-9B. Unless otherwise specified, we set $K=10$ and use reliability weights in $[0.8,1.2]$. For reward aggregation, the IF/AP/CP weights are set to $(0.4,0.35,0.25)$, with SC/PQ exponents of $0.8/0.2$. Implementation details are provided in Appendix~\ref{app:implementation}.

\paragraph{Baselines and Evaluation Protocol.}
We evaluate the reward modeling performance on MMRB2~\citep{hu2026mmrb2}, MER-Bench~\citep{long2026spatialreward}, and EditReward-Bench~\citep{luo2026editscore}. We compare \method\ with closed-source visual judges (GPT-4.1, Gemini-3.1-Pro, and Gemini-3.0-Flash), untrained reward-model backbones, and specialized reward models including EditReward~\citep{wu2026editreward}, EditScore-8B~\citep{luo2026editscore}, and SpatialReward~\citep{long2026spatialreward}. 
We also conduct a controlled comparison on Qwen3.5-9B against GRPO~\citep{shao2024deepseekmath} and Token-level GRPO under the same experimental settings. More details on the evaluation protocol are provided in Appendix~\ref{app:evaluation}.

\begin{table}[!t]
\centering
\caption{Reward benchmark accuracy (\%, $\uparrow$).
Submetric results for EditReward, EditScore-8B, and SpatialReward are reproduced from SpatialReward~\citep{long2026spatialreward}. Bold marks the best result among the compared open-source models in each column.}
\label{tab:main}
\fontsize{7pt}{7.6pt}\selectfont
\setlength{\tabcolsep}{3pt}
\resizebox{\linewidth}{!}{%
\begin{tabular}{
l
r r >{\columncolor{gray!15}}r
r r r >{\columncolor{gray!15}}r
r r >{\columncolor{gray!15}}r
}
\toprule
& \multicolumn{3}{c}{\textbf{MMRB2}}
& \multicolumn{4}{c}{\textbf{MER-Bench}}
& \multicolumn{3}{c}{\textbf{EditReward-Bench}} \\
\cmidrule(lr){2-4}\cmidrule(lr){5-8}\cmidrule(lr){9-11}

\textbf{Model}
& \textbf{Single} & \textbf{Multi} & \textbf{Overall}
& \textbf{2 pair} & \textbf{3 pair} & \textbf{4 pair} & \textbf{Overall}
& \textbf{PF} & \textbf{CS} & \textbf{Overall} \\
\midrule

\color{gray}GPT-4.1
&\color{gray}56.33&\color{gray}57.87&\color{gray}56.60
&\color{gray}79.50&\color{gray}42.50&\color{gray}18.50&\color{gray}46.83
&\color{gray}70.00&\color{gray}65.20&\color{gray}73.40\\

\color{gray}Gemini-3.1-Pro
&\color{gray}65.20&\color{gray}60.11&\color{gray}64.29
&\color{gray}79.00&\color{gray}60.50&\color{gray}24.00&\color{gray}54.50
&\color{gray}66.30&\color{gray}75.70&\color{gray}79.80\\

\color{gray}Gemini-3.0-Flash
&\color{gray}66.71&\color{gray}58.64&\color{gray}65.34
&\color{gray}82.50&\color{gray}63.00&\color{gray}26.00&\color{gray}57.17
&\color{gray}68.70&\color{gray}77.90&\color{gray}80.90\\

\midrule

Qwen3-VL-8B-Instruct
&63.02&57.87&62.10
&73.00&42.00&16.00&43.67
&77.20&70.90&75.80\\

Qwen3.5-4B
&57.79&54.49&57.20
&75.50&36.00&15.50&42.33
&74.40&69.60&70.50\\

Qwen3.5-9B
&63.14&58.99&62.40
&76.00&46.50&19.00&47.17
&83.10&75.60&78.20\\

\midrule

EditReward
&67.20&59.00&65.70
&70.00&49.00&15.50&44.83
&83.20&-&79.20\\

EditScore-8B
&57.90&52.80&57.00
&61.00&34.00&10.00&35.00
&60.80&59.40&69.00\\

SpatialReward
&67.10&60.80&66.10
&78.00&49.50&21.50&49.67
&68.30&67.20&80.30\\

\midrule

GRPO (Qwen3.5-9B)
&69.59&\textbf{70.22}&69.70
&81.50&58.50&\textbf{29.00}&56.33
&85.40&85.50&81.10\\

Token-level GRPO (Qwen3.5-9B)
&70.19&69.10&70.00
&80.50&59.00&27.00&55.50
&85.30&84.30&81.70\\

\midrule

\rowcolor{methodblue}
\textbf{\method\ (Qwen3-VL-8B)}
&69.22&60.67&67.70
&\textbf{83.00}&50.50&25.00&52.83
&\textbf{85.50}&81.20&78.90\\

\rowcolor{methodblue}
\textbf{\method\ (Qwen3.5-4B)}
&\textbf{73.48}&68.54&\textbf{72.60}
&82.00&53.50&20.00&51.83
&84.50&\textbf{85.60}&82.70\\

\rowcolor{methodblue}
\textbf{\method\ (Qwen3.5-9B)}
&73.36&67.42&72.30
&82.00&\textbf{60.50}&27.50&\textbf{56.67}
&\textbf{85.50}&84.10&\textbf{82.90}\\

\bottomrule
\end{tabular}}
\end{table}

\paragraph{Comparison with Baselines.}
As shown in Table~\ref{tab:main}, \method\ consistently improves over the corresponding backbone models and achieves strong performance across all three benchmarks. In particular, \method\ (Qwen3.5-9B) reaches 72.30\% on MMRB2, 56.67\% on MER-Bench, and 82.90\% on EditReward-Bench, outperforming SpatialReward by 6.20, 7.00, and 2.60 percentage points, respectively. It also exceeds GPT-4.1 and Gemini-3.1-Pro on all three aggregate metrics, while outperforming Gemini-3.0-Flash on MMRB2 and EditReward-Bench and remaining competitive on MER-Bench (56.67\% vs.\ 57.17\%). The gains are consistent across model scales: the 4B variant achieves the highest MMRB2 overall accuracy of 72.60\%, improving its backbone by 15.40 percentage points. 
Under the same Qwen3.5-9B backbone and experimental settings, \method\ also outperforms GRPO and Token-level GRPO (see Appendix~\ref{app:en:grpo} for implementation details), improving MMRB2 overall accuracy by 2.60 and 2.30 percentage points and EditReward-Bench overall accuracy by 1.80 and 1.20 percentage points, respectively. These results demonstrate the effectiveness of CR-OPSD across different backbone families, model scales, and training strategies.

\subsection{Application in Online RL}
\paragraph{Experimental Setup.}
We conduct online RL experiments on Qwen-Image-Edit~\citep{wu2025qwenimage} using DiffusionNFT~\citep{zheng2025diffusionnft}, with training samples drawn from CheckDistill-40K. We evaluate the resulting model on ImgEdit-Bench~\citep{ye2025imgedit}, REDEdit-Bench~\citep{team2026firered}, and the English subset of GEdit-Bench (GEdit-EN)~\citep{liu2025step1xedit}. We compare CheckDistill-guided RL against the original Qwen-Image-Edit and the open-source UniWorld-Qwen-Image-Edit model~\citep{li2025uniworldv2}.

\begin{figure}[!t]
\centering
\includegraphics[width=\linewidth]{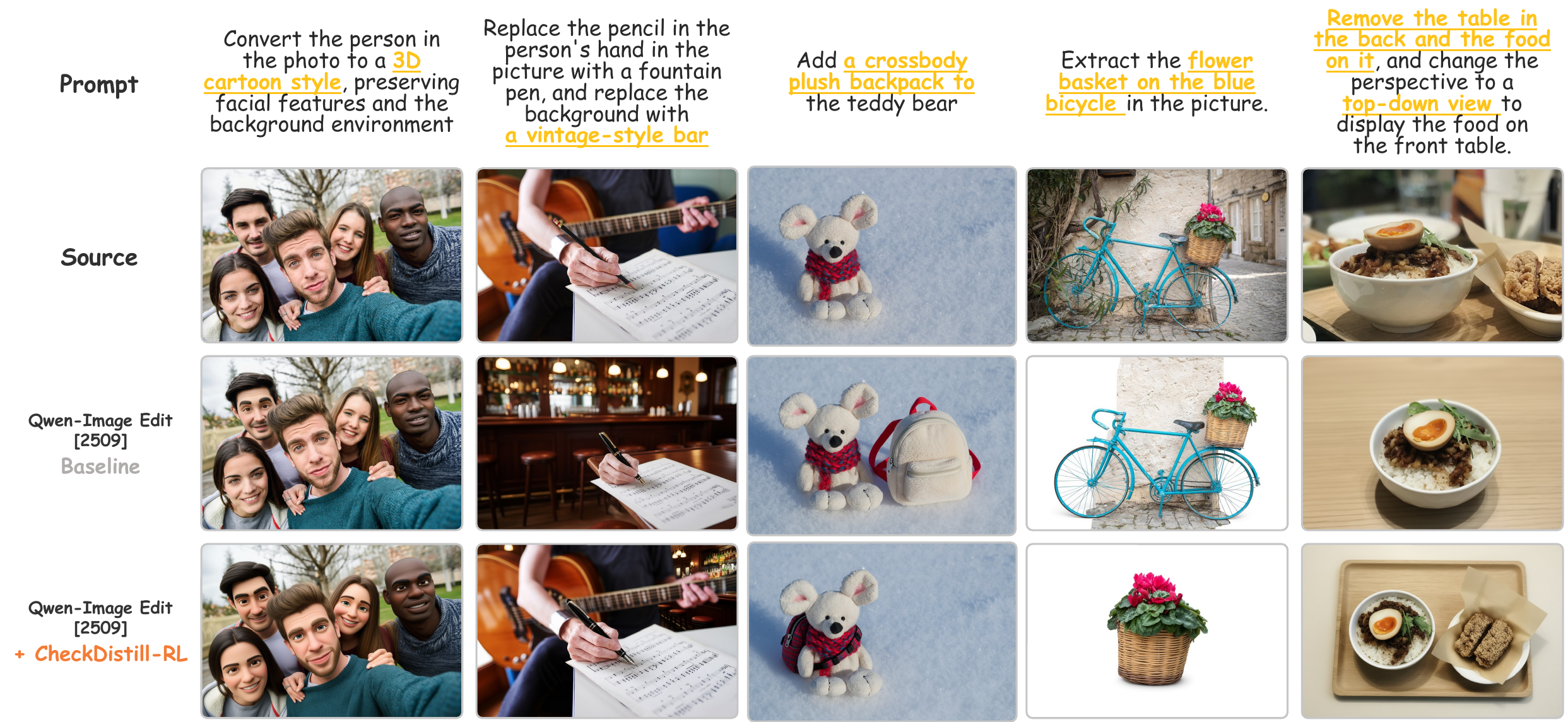}
\caption{Qualitative Comparison Before and After CheckDistill-based Reward Training.}
\label{fig:online_qualitative}
\end{figure}

\paragraph{Editing Performance.}
\begin{wraptable}{r}{0.5\columnwidth}
\vspace{-10pt}
\centering
\captionsetup{skip=1.5pt}
\caption{Image-editing performance ($\uparrow$).}
\label{tab:online}
\fontsize{7.6pt}{8pt}\selectfont
\setlength{\tabcolsep}{0.8pt}
\renewcommand{\arraystretch}{1.05}
\begin{tabular}{lccc}
\toprule
\textbf{Model}
& \makecell{\textbf{ImgEdit}\\\textbf{Bench}}
& \makecell{\textbf{REDEdit}\\\textbf{Bench}}
& \textbf{GEdit-EN} \\
\midrule

Qwen-Image-Edit
& 4.29
& 4.10
& 8.18 \\

UniWorld-Qwen-Image-Edit
& 4.30
& 4.06
& 8.12 \\

\makecell[l]{\textbf{+ DiffusionNFT}\\\textbf{(CheckDistill-9B)}}
& \textbf{4.46}
& \textbf{4.26}
& \textbf{8.28} \\

\bottomrule
\end{tabular}
\vspace{-8pt}
\end{wraptable}
As shown in Table~\ref{tab:online}, RL guided by \method\ consistently improves Qwen-Image-Edit across all three benchmarks, increasing ImgEdit-Bench from 4.29 to 4.46, REDEdit-Bench from 4.10 to 4.26, and GEdit-EN from 8.18 to 8.28. It also outperforms UniWorld-Qwen-Image-Edit, improving the three benchmarks from 4.30, 4.06, and 8.12 to 4.46, 4.26, and 8.28, respectively. Qualitative comparisons in Figure~\ref{fig:online_qualitative} further show that \method-guided RL better follows editing instructions while maintaining stronger consistency with the source image. These results demonstrate the effectiveness of CheckDistill as a reward model for improving downstream image-editing performance through online RL.

\begin{wrapfigure}{R}{0.44\linewidth}
    \vspace{-15pt}
    \centering
    \includegraphics[width=\linewidth]
    {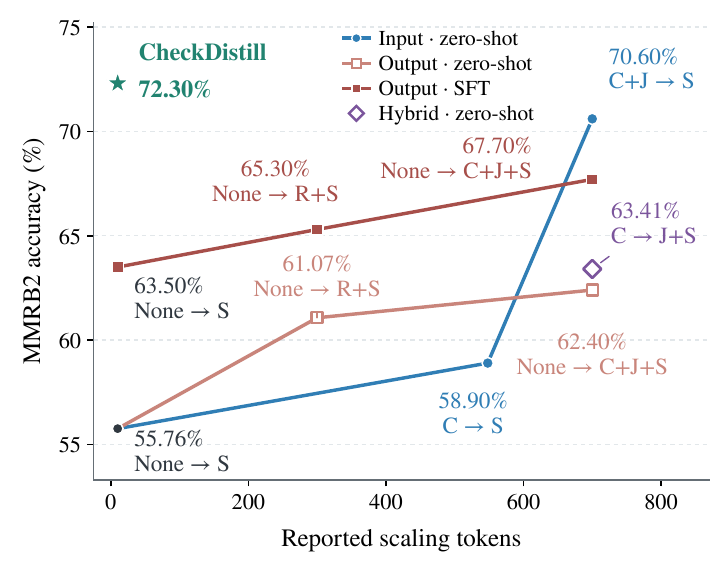}

    \par\vspace{-6pt} 
    \caption{Input, output, and hybrid scaling on MMRB2 (Qwen3.5-9B). Labels indicate additional input context $\rightarrow$ generated output. C: checklist; J: checklist-guided judgments; R: rationale; S: score. ``None'' denotes no additional input context.}
    \label{fig:input_output_scaling}
    \vspace{-10pt}
\end{wrapfigure}

\subsection{Ablation Study and Analysis}

\paragraph{Analysis of Verification Scaling and Efficient Scoring.}
We evaluate input and output scaling on MMRB2 using Qwen3.5-9B, measuring accuracy and output token counts. In Figure~\ref{fig:input_output_scaling}, SFT denotes supervised fine-tuning on CheckDistill-40K, while zero-shot refers to evaluation without task-specific fine-tuning.

\label{sec:scaling_analysis}

\textbf{Instance-specific verification evidence improves reward judgement through input and output scaling.}
Supplying a checklist improves zero-shot direct-scoring accuracy from 55.76\% to 58.90\%, while adding checklist-guided judgements raises it to 70.60\%, with about 10 output tokens.
Generating evaluation content also helps: a generic rationale achieves 61.07\%, generating a checklist and its guided judgement reaches 62.40\%, and supplying the checklist before generating the judgement achieves 63.41\%.
These gains support grounding reward assessment in instance-specific verification requirements.
In particular, \textbf{the strong input-scaling result demonstrates the value of evidence-enriched teacher contexts}, motivating Dynamic Checklist Scaling in CheckDistill.

\textbf{CheckDistill retains the benefits of evidence-supported evaluation in a compact, direct-scoring model.}
SFT achieves 63.50\% accuracy with direct scoring and 67.70\% when generating a checklist and judgement.
CheckDistill reaches 72.30\% with about 10 output tokens, exceeding these configurations by 8.80 and 4.60 percentage points and surpassing the 70.60\% zero-shot result with supplied verification evidence.
This supports Distribution-Aligned Context Compression: transferring evidence-enhanced scoring to a student requiring neither verification contexts nor lengthy reasoning at inference time.
Compared with the approximately 700-token SFT reasoning configuration, \textbf{CheckDistill achieves higher accuracy with approximately 1/70 of the output tokens}, demonstrating a favorable accuracy--decoding-cost trade-off.

\paragraph{Privileged-Information Design.}
We ablate the privileged information provided to the teacher by progressively introducing checklist $C$, checklist judgments $(C,J)$, dimension-wise rationales $(C,J,A)$, and dimension scores $(C,J,A,\mathbf g)$, while keeping the SFT initialization, training data, student inputs, and optimization settings unchanged. As shown in Table~\ref{tab:ablation}, $(C,J,A)$ achieves the best overall performance across all three benchmarks, reaching 72.30\% on MMRB2, 56.67\% on MER-Bench, and 82.90\% on EditReward-Bench.

However, further exposing dimension scores to the teacher, i.e., directly providing the ground-truth targets as in OPSD, does not yield additional gains: $(C,J,A)$ outperforms $(C,J,A,\mathbf g)$ by 2.20, 0.84, and 1.60 points on MMRB2, MER-Bench, and EditReward-Bench, respectively. These results demonstrate the effectiveness of checklist-based privileged information, where checklist judgments and dimension-wise rationales further enrich the supervision without directly exposing score labels.

\paragraph{CR-OPSD Component Ablation.}
We ablate the key design choices of CR-OPSD while keeping the privileged information $(C,J,A)$, initialization, training data, and optimization settings fixed. Specifically, \textit{w/o Reliability Weighting} removes reliability modulation by setting $c_t=1$ for all positions; \textit{Linear Reliability Mapping} replaces the cosine mapping in Eq.~\ref{eq:score_reliability} with a linear mapping; and \textit{Argmax-based Reliability} replaces the expected teacher score with the teacher's maximum-probability score token for reliability estimation.

As shown in Table~\ref{tab:ablation}, CR-OPSD consistently achieves the best overall performance across all three benchmarks. Compared with removing reliability weighting, it improves MMRB2, MER-Bench, and EditReward-Bench by 2.30, 1.67, and 0.80 percentage points, respectively, demonstrating the benefit of modulating privileged supervision according to teacher reliability. The cosine mapping further improves over its linear counterpart by 1.00, 0.50, and 0.60 points, while using the expected teacher score instead of the argmax score yields gains of 1.70, 1.17, and 0.30 points. These results validate both reliability-aware weighting and the use of distributional score information for more robust privileged on-policy self-distillation.

\begin{table}[!t]
\vspace{-6pt}
\centering
\caption{Ablation studies on privileged information and CR-OPSD design choices (\%, $\uparrow$).}
\label{tab:ablation}

\vspace{-6pt}
\begin{minipage}[t]{0.52\linewidth}
\centering
\textbf{(a) Privileged-information design}\\[3pt]

\fontsize{7.6pt}{8pt}\selectfont
\setlength{\tabcolsep}{1.5pt}
\renewcommand{\arraystretch}{1.0}

\begin{tabularx}{\linewidth}{
>{\raggedright\arraybackslash}X
ccc
}
\toprule
\textbf{Privileged Information}
& \textbf{MMRB2}
& \makecell{\textbf{MER-}\\\textbf{Bench}}
& \makecell{\textbf{EditReward-}\\\textbf{Bench}} \\
\midrule

Checklist $C$
& 72.10
& 54.83
& 82.10 \\

+ Checklist Judgment $J$
& 72.00
& 56.00
& 82.20 \\

\rowcolor{methodblue}
\textbf{+ Dimension-wise Rationale $A$}
& \textbf{72.30}
& \textbf{56.67}
& \textbf{82.90} \\

+ Dimension Scores $\mathbf{g}$
& 70.10
& 55.83
& 81.30 \\

\bottomrule
\end{tabularx}
\end{minipage}
\hfill
\begin{minipage}[t]{0.47\linewidth}
\centering
\textbf{(b) CR-OPSD design choices}\\[3pt]

\fontsize{7.6pt}{8pt}\selectfont
\setlength{\tabcolsep}{1.5pt}
\renewcommand{\arraystretch}{1.0}

\begin{tabularx}{\linewidth}{
>{\raggedright\arraybackslash}X
ccc
}
\toprule
\textbf{Variant}
& \textbf{MMRB2}
& \makecell{\textbf{MER-}\\\textbf{Bench}}
& \makecell{\textbf{EditReward-}\\\textbf{Bench}} \\
\midrule

w/o Reliability Weighting
& 70.00
& 55.00
& 82.10 \\

Linear Reliability Mapping
& 71.30
& 56.17
& 82.30 \\

Argmax-based Reliability
& 70.60
& 55.50
& 82.60 \\

\rowcolor{methodblue}
\textbf{CR-OPSD (Ours)}
& \textbf{72.30}
& \textbf{56.67}
& \textbf{82.90} \\

\bottomrule
\end{tabularx}
\end{minipage}

\end{table}

\section{Conclusion}
We present CheckDistill, an efficient reward modeling framework for instruction-guided image editing. It combines instance-specific dynamic checklists for fine-grained verification with CR-OPSD to distill evidence-supported scoring behavior into a direct-scoring model that requires neither checklist inputs nor explicit reasoning at inference time. Experiments demonstrate strong reward modeling performance and improved image editing through reward-guided reinforcement learning using DiffusionNFT. Together, these findings show that scaling instance-specific verification improves reward quality, while reliability-aware distillation enables efficient reward scoring for image editing at scale.

\bibliographystyle{unsrtnat}
\bibliography{references}

\clearpage
\appendix

\section{Implementation Details}
\label{app:implementation}

\subsection{Stage I: SFT Implementation Details}

We perform supervised fine-tuning (SFT) using \textbf{ms-swift} on 8 NVIDIA A100 GPUs. The training hyperparameters are summarized in Table~\ref{tab:sft_settings}.

\begin{table}[htbp]
    \centering
    \caption{Hyperparameters for Stage-I supervised fine-tuning.}
    \label{tab:sft_settings}
    \begin{tabular}{ll}
        \hline
        \textbf{Hyperparameter} & \textbf{Value} \\
        \hline
        Optimizer & AdamW ($\beta_1=0.9$, $\beta_2=0.999$) \\
        Learning Rate & $1\times10^{-5}$ \\
        Learning Rate Scheduler & Linear \\
        Warmup Ratio & 0.1 \\
        Per-Device Batch Size & 1 \\
        Gradient Accumulation Steps & 4 \\
        Effective Global Batch Size & 32 \\
        DeepSpeed Stage & ZeRO-2 \\
        Maximum Sequence Length & 32,768 \\
        Maximum Image Pixels & 802,816 \\
        \hline
    \end{tabular}
\end{table}

\subsection{Stage II: CR-OPSD Implementation Details}
We implement CR-OPSD using the \textbf{verl} framework. Both the student and teacher are initialized from the Stage-I SFT checkpoint, and the teacher is maintained as an exponential moving average (EMA) of the student with an update rate of $0.01$. The training hyperparameters are summarized in Table~\ref{tab:self_distillation_settings}. To accommodate samples with many reference images, we set the maximum image pixel budget to $802{,}816$ pixels per image and reduce it to $262{,}144$ pixels when more than five reference images are provided.

\begin{table}[!ht]
    \centering
    \caption{Hyperparameters for Stage-II CR-OPSD.}
    \label{tab:self_distillation_settings}
    \begin{tabular}{ll}
        \toprule
        \textbf{Hyperparameter} & \textbf{Value} \\
        \midrule
        Global Batch Size & 128 \\
        Mini-Batch Size & 128 \\
        Rollouts per Prompt & 4 \\
        Log-Probability Top-$K$ & 10 \\
        Maximum Prompt Length & 32,768 \\
        Maximum Response Length & 128 \\
        Learning Rate & $1\times10^{-6}$ \\
        Learning Rate Scheduler & Cosine \\
        Warmup Ratio & 0.03 \\
        Training Epochs & 1 \\
        Training Temperature & 1.0 \\
        Teacher Temperature & 1.0 \\
        Teacher EMA Update Rate & 0.01 \\
        KL Coefficient & 0.0 \\
        Maximum Image Pixels & 802,816 \\
        \bottomrule
    \end{tabular}
\end{table}

\subsection{DiffusionNFT Implementation Details}

We use CheckDistill to guide RL post-training of the image editing model with DiffusionNFT~\citep{zheng2025diffusionnft}. For each editing condition $h$, the old policy generates $G$ candidates. We evaluate each candidate to obtain its reward $R_k=R(x_k)$ using Eq.~\ref{eq:expected_score} and ~\ref{eq:final_reward}, which is then used to determine its positive-branch weight relative to the group mean:
\begin{equation}
r_k
=
\frac{1}{2}
+
\frac{1}{2}
\operatorname{clip}\!\left(
\frac{R_k-\bar{R}_h}{Z_c},
-1,1
\right),
\qquad
\bar{R}_h
=
\frac{1}{G}\sum_{k=1}^{G}R_k,
\label{eq:nft_weight}
\end{equation}
where $Z_c>0$ is the reward normalization scale and the negative-branch weight is $1-r_k$.

To suppress updates from nearly saturated candidate groups, we follow the Low-STD Group Filtering strategy of UniWorld-V2~\citep{li2025uniworldv2}.
\begin{equation}
\sigma_h
=
\sqrt{\frac{1}{G}\sum_{k=1}^{G}(R_k-\bar{R}_h)^2},
\qquad
M_h
=
1-\mathbf{1}\!\left\{
\bar{R}_h>\eta_{\mu}
\land
\sigma_h<\eta_{\mathrm{std}}
\right\}.
\label{eq:group_filter}
\end{equation}
Here, $\bar{R}_h$ and $\sigma_h$ denote the mean and standard deviation of the rewards in group $h$, respectively; $\eta_{\mu}$ and $\eta_{\mathrm{std}}$ are the corresponding filtering thresholds. Groups with high mean reward and low standard deviation are excluded from optimization by setting $M_h=0$.

Following DiffusionNFT, let $v_\omega^{+}$ and $v_\omega^{-}$ denote the positive and negative policy branches, and $v_k$ the target velocity of candidate $k$. The optimization objective is
\begin{equation}
\mathcal{L}_{\mathrm{NFT}}(\omega)
=
\mathbb{E}\!\left[
M_h\frac{1}{G}\sum_{k=1}^{G}
\left(
r_k
\left\|
v_\omega^{+}-v_k
\right\|_2^2
+
(1-r_k)
\left\|
v_\omega^{-}-v_k
\right\|_2^2
\right)
\right].
\label{eq:nft_objective}
\end{equation}

We enable classifier-free guidance (CFG) during rollout but disable it during policy optimization to reduce training overhead.
Training is performed with LoRA, with the hyperparameters summarized in Table~\ref{tab:diffusionnft_settings}.

\begin{table}[!ht]
    \centering
    \caption{Hyperparameters for DiffusionNFT training.}
    \label{tab:diffusionnft_settings}
    \begin{tabular}{ll}
        \toprule
        \textbf{Hyperparameter} & \textbf{Value} \\
        \midrule
        Guidance Scale & 4.0 \\
        Training Sampling Steps & 16 \\
        Training Sampler & DPM2 \\
        Evaluation Sampling Steps & 40 \\
        Noise Level & 0.7 \\
        Learning Rate & $3\times10^{-4}$ \\
        NFT Mixing Coefficient ($\beta$) & $1\times10^{-4}$ \\
        Per-GPU Batch Size & 1 \\
        Group Size & 8 \\
        Number of Groups & 48 \\
        Group Mean Filtering Threshold & 8 \\
        Group Std.\ Filtering Threshold & 0.025 \\
        LoRA Rank & 32 \\
        LoRA Alpha & 64 \\
        Timestep Fraction & 0.99 \\
        Maximum Gradient Norm & 1.0 \\
        Optimizer & AdamW ($\beta_1=0.9$, $\beta_2=0.999$) \\
        Weight Decay & $1\times10^{-4}$ \\
        GPUs & \makecell[l]{$40\times$ NVIDIA A100 \\
        (32 for training, 8 for reward model deployment)} \\
        \bottomrule
    \end{tabular}
\end{table}

\subsection{Reward Benchmark Evaluation Details}
\label{app:evaluation}

We evaluate reward prediction on MMRB2, MER-Bench, and EditReward-Bench, none of which overlaps with the data used for CheckDistill.
MMRB2 reports single-reference, multi-reference, and overall accuracy, where the overall accuracy is computed as a sample-count-weighted average of the single- and multi-reference subsets rather than their arithmetic mean. MER-Bench reports ranking accuracy under the 2-pair, 3-pair, and 4-pair settings, with their arithmetic mean used as the overall metric. EditReward-Bench evaluates prompt following (PF), consistency (CS), and overall preference accuracy. For \method\, PF is given by the instruction-following score, while CS is computed from the attribute-preservation (AP) and context-preservation (CP) scores using weights of 0.35 and 0.25, respectively, consistent with their weights in the overall reward. The final scalar reward follows Eq.~\ref{eq:expected_score} and ~\ref{eq:final_reward}. Except for specialized reward models with their own inference formats, all evaluated models use the same system prompt. Each candidate is scored independently in a pointwise manner, after which pairwise comparisons and multi-candidate rankings are obtained by comparing the resulting scalar scores.

\section{Implementation Details of GRPO and Token-Level GRPO}
\label{app:en:grpo}

GRPO and Token-Level GRPO are initialized from the same SFT checkpoint as CheckDistill. For each input $x$, they sample $G$ scoring responses $\{y_i\}_{i=1}^{G}$. 
Let $S_{i,\delta}$ denote the numerical score represented by the sampled score token for dimension $\delta$ in response $i$, and let $g_\delta$ be the corresponding ground-truth label, where $\delta\in\mathcal{D}=\mathcal{D}_{\mathrm{SC}}\cup\{\mathrm{PQ}\}$. Following the reward aggregation rule in the main text, the predicted and label-derived aggregate scores are
\begin{equation}
\begin{aligned}
R_i &=
\left(
\sum_{\delta\in\mathcal{D}_{\mathrm{SC}}}
w_\delta S_{i,\delta}
\right)^\lambda
S_{i,\mathrm{PQ}}^{\,1-\lambda},\\
R_g &=
\left(
\sum_{\delta\in\mathcal{D}_{\mathrm{SC}}}
w_\delta g_\delta
\right)^\lambda
g_{\mathrm{PQ}}^{\,1-\lambda}.
\end{aligned}
\end{equation}

\paragraph{GRPO: Sequence-Level Advantage.}
GRPO constructs a response-level reward from the discrepancy between the predicted and label-derived aggregate scores:
\begin{equation}
r_i
=
1-\frac{|R_i-R_g|}{9},
\qquad
A_{i,t}^{\mathrm{GRPO}}
=
r_i-\frac{1}{G}\sum_{j=1}^{G}r_j.
\end{equation}
The same group-centered advantage is applied to all tokens in response $i$.

\paragraph{Token-Level GRPO: Dimension-Level Advantage.}
Token-Level GRPO instead computes a separate reward for each scoring dimension:
\begin{equation}
\begin{aligned}
r_{i,\delta}
&=
1-\frac{|S_{i,\delta}-g_\delta|}{9},\\
A_{i,t}^{\mathrm{Token}}
&=
r_{i,\delta(t)}
-
\frac{1}{G}
\sum_{j=1}^{G}
r_{j,\delta(t)},
\end{aligned}
\end{equation}
where $\delta(t)$ denotes the scoring dimension associated with score-token position $t$. The dimension-specific advantage is applied only at the four score-token positions, while non-score positions receive zero policy advantage. Both methods use group-mean centering without standard-deviation normalization.

\paragraph{Training Objective.}
GRPO and Token-Level GRPO share the same clipped policy objective and differ only in the definition of $A_{i,t}$. Let
\begin{equation}
\rho_{i,t}(\theta)
=
\frac{
p_\theta(y_{i,t}\mid x,y_{i,<t})
}{
p_{\theta_{\mathrm{old}}}(y_{i,t}\mid x,y_{i,<t})
}.
\end{equation}
The training loss is
\begin{equation}
\begin{aligned}
\mathcal{L}_{\mathrm{base}}(\theta;A)
=
-\mathbb{E}\Bigg[
\frac{1}{\sum_i |y_i|}
\sum_{i=1}^{G}
\sum_{t=1}^{|y_i|}
\min\Big(
&\rho_{i,t}(\theta)A_{i,t},\\
&\operatorname{clip}
\big(
\rho_{i,t}(\theta),
1-\epsilon_l,
1+\epsilon_h
\big)
A_{i,t}
\Big)
\Bigg]
+
\beta\mathcal{L}_{\mathrm{KL}}.
\end{aligned}
\end{equation}
GRPO and Token-Level GRPO use $A^{\mathrm{GRPO}}$ and $A^{\mathrm{Token}}$, respectively. Here, $\mathcal{L}_{\mathrm{KL}}$ regularizes the policy against the frozen SFT reference model, $\epsilon_l$ and $\epsilon_h$ denote the clipping ranges, and $\beta$ is the KL coefficient.

\paragraph{Training Settings.}
We train GRPO and Token-level GRPO on CheckDistill-40K using the \textbf{verl} framework under the same training configuration. The shared hyperparameters are summarized in Table~\ref{tab:grpo_settings}.

\begin{table}[htbp]
    \centering
    \caption{Hyperparameters for GRPO and Token-level GRPO.}
    \label{tab:grpo_settings}
    \begin{tabular}{ll}
        \toprule
        \textbf{Hyperparameter} & \textbf{Value} \\
        \midrule
        Training Temperature & 1.0 \\
        Global Batch Size & 128 \\
        Mini-Batch Size & 128 \\
        Rollouts per Prompt & 8 \\
        Maximum Prompt Length & 16,384 \\
        Maximum Response Length & 128 \\
        Learning Rate & $1\times10^{-6}$ \\
        Learning Rate Scheduler & Cosine \\
        Warmup Ratio & 0.03 \\
        KL Coefficient & 0.001 \\
        PPO Clip Ratio & $\epsilon_l=0.2,\ \epsilon_h=0.3$ \\
        Maximum Image Pixels & 802,816 \\
        \bottomrule
    \end{tabular}
\end{table}

\section{Alternative Reverse-KL Formulation}
\subsection{Preliminaries}
\label{sec:preliminaries}
\paragraph{On-Policy Self-Distillation (OPSD).}
OPSD~\citep{zhao2026opsd}  enables a single model to act as both student and teacher by conditioning on different contexts, where the teacher additionally observes privileged information and provides token-level supervision along student-generated trajectories. Given a task input $x$, the student samples a response $y \sim p_\theta(\cdot \mid x)$. With privileged information $z$ available to the teacher, the student and teacher predictive distributions at the same response prefix $y_{<t}$ are defined as
\begin{equation}
p_S(y_t \mid x, y_{<t}) = p_\theta(y_t \mid x, y_{<t}),
\qquad
p_T(y_t \mid x, z, y_{<t}) = p_\theta(y_t \mid x, z, y_{<t}).
\end{equation}

From a policy-gradient perspective, the token-level distillation signal for a sampled token $y_t$ can be expressed as
\begin{equation}
A_t =
\operatorname{sg}\!\left[
\log
\frac{
p_T(y_t \mid x, z, y_{<t})
}{
p_S(y_t \mid x, y_{<t})
}
\right],
\end{equation}
where $\operatorname{sg}[\cdot]$ denotes the stop-gradient operator. A positive $A_t$ reinforces tokens assigned higher likelihood by the teacher than by the student, whereas a negative $A_t$ suppresses them. The corresponding on-policy policy-gradient surrogate objective is
\begin{equation}
\mathcal{L}_{\mathrm{OPSD}}(\theta)
=
-\mathbb{E}\!\left[
\frac{1}{|y|}
\sum_t
A_t
\log p_S(y_t \mid x, y_{<t})
\right].
\end{equation}

\label{app:en:rkl}

\subsection{Alternative Objective}

Our method admits the following alternative reverse-KL objective:
\begin{equation}
\begin{aligned}
\mathcal L_{\mathrm{wRKL}}^{\mathrm{TopK}}(\theta)
=
\mathbb E\Bigg[\frac1{|y|}\sum_t c_t
\sum_{y_t\in\mathcal T_t}\Bigg(
&p_S(y_t\mid x,y_{<t})
\log\frac{p_S(y_t\mid x,y_{<t})}{p_T(y_t\mid x,z,y_{<t})}\\
&-p_S(y_t\mid x,y_{<t})
+p_T(y_t\mid x,z,y_{<t})
\Bigg)\Bigg].
\end{aligned}
\label{eq:app:en:rkl}
\end{equation}
The term $-p_S+p_T$ corrects the probability-mass difference on the truncated support. When the candidate set covers the entire vocabulary, this term sums to zero, and Eq.~\ref{eq:app:en:rkl} reduces to reliability-weighted reverse KL.

\subsection{Connection to the Policy-Gradient Objective}

We treat the response prefixes, candidate set $\mathcal T_t$, teacher distribution, and $c_t$ as fixed when differentiating with respect to $\theta$. For each candidate, the correction term $-p_S$ cancels the extra $+1$ term arising from differentiating $p_S\log(p_S/p_T)$:
\begin{equation}
\begin{aligned}
\nabla_\theta\Bigg[
&p_S(y_t\mid x,y_{<t})
\log\frac{p_S(y_t\mid x,y_{<t})}{p_T(y_t\mid x,z,y_{<t})}
-p_S(y_t\mid x,y_{<t})
+p_T(y_t\mid x,z,y_{<t})
\Bigg]\\
&=
\log\frac{p_S(y_t\mid x,y_{<t})}{p_T(y_t\mid x,z,y_{<t})}
\nabla_\theta p_S(y_t\mid x,y_{<t}).
\end{aligned}
\label{eq:c2}
\end{equation}

Substituting Eq.~\ref{eq:c2} into Eq.~\ref{eq:app:en:rkl} and using
$\nabla_\theta p_S=p_S\nabla_\theta\log p_S$, we obtain
\begin{equation}
\begin{aligned}
\nabla_\theta\mathcal L_{\mathrm{wRKL}}^{\mathrm{TopK}}
&=
\mathbb E\!\left[
\frac1{|y|}
\sum_t
\sum_{y_t\in\mathcal T_t}
c_t
\log\frac{p_S(y_t\mid x,y_{<t})}{p_T(y_t\mid x,z,y_{<t})}
\nabla_\theta p_S(y_t\mid x,y_{<t})
\right]\\
&=
-\mathbb E\!\left[
\frac1{|y|}
\sum_t
\sum_{y_t\in\mathcal T_t}
p_S(y_t\mid x,y_{<t})\,c_t
\log\frac{p_T(y_t\mid x,z,y_{<t})}{p_S(y_t\mid x,y_{<t})}
\nabla_\theta\log p_S(y_t\mid x,y_{<t})
\right].
\end{aligned}
\end{equation}

Defining
$\hat A_t(y_t)=\operatorname{sg}\!\left[
c_t p_S(y_t\mid x,y_{<t})
\log\frac{p_T(y_t\mid x,z,y_{<t})}
{p_S(y_t\mid x,y_{<t})}
\right]$,
we obtain the following policy-gradient surrogate with the same gradient:
\begin{equation}
\begin{aligned}
\mathcal L_{\mathrm{PG\text{-}RKL}}(\theta)
=
-\mathbb E\!\left[
\frac1{|y|}
\sum_t
\sum_{y_t\in\mathcal T_t}
\hat A_t(y_t)
\log p_S(y_t\mid x,y_{<t})
\right].
\end{aligned}
\label{eq:c4}
\end{equation}

Eq.~\ref{eq:c4} matches the training objective in the main text and is gradient-equivalent to Eq.~\ref{eq:app:en:rkl} at the parameter point where the advantages are computed.

\section{Background}

\paragraph{Multimodal Instruction-Guided Image Editing.}
Instruction-guided image editing requires models to execute requested transformations while preserving unrelated visual content. Recent open-source and open-weight models, exemplified by Qwen-Image and FLUX.1 Kontext, have substantially expanded the capabilities of language-driven image editing~[\citealp{wu2025qwenimage,blackforestlabs2025flux,hu2026swiftimage,longcat2025image,wu2025omnigen2,liu2025step1xedit,team2026firered}]. Capability-centric data design further supports the joint development of generation and editing through coordinated supervision and training curricula~[\citealp{wang2026capabilitydata}]. Meanwhile, commercial systems such as GPT Image and Gemini bring image generation and editing into interactive creative workflows~[\citealp{openai2026gptimage,google2026geminiimage}], while virtual try-on research extends these capabilities toward fashion applications that require faithful preservation of garment appearance and person identity~[\citealp{chen2026tstarstryon,choi2024idmvton,chong2024catvton}]. To assess these increasingly diverse capabilities, benchmarks including GEdit-Bench, ImgEdit-Bench, REDEdit-Bench, and CPI-Bench collectively broaden evaluation across editing tasks and practical usage scenarios~[\citealp{liu2025step1xedit,ye2025imgedit,team2026firered,zhou2026cpibench}]. These developments motivate alignment methods that jointly account for instruction fulfillment, content preservation, and perceptual quality.

\paragraph{Reinforcement Learning for Image Editing.}
RL provides a post-training mechanism for optimizing visual generators against feedback beyond supervised reconstruction objectives. DDPO and DPOK establish policy-gradient approaches for diffusion models~\citep{black2024ddpo,fan2023dpok}; Flow-GRPO and DanceGRPO extend group-relative optimization to flow-based and broader visual generation settings~\citep{liu2025flowgrpo,xue2025dancegrpo}, while DiffusionNFT formulates online reward-driven updates through the forward process~\citep{zheng2025diffusionnft}. Building on these foundations, editing-specific methods increasingly emphasize the design of reward feedback. HIVE learns human preferences to guide instructional editing~\citep{zhang2024hive}. UniWorld-V2 combines DiffusionNFT with MLLM logit-based feedback and group filtering to stabilize editing optimization~\citep{li2025uniworldv2}. Edit-R1 trains a principle-based reasoning verifier and uses its rewards to optimize editing models with GRPO~\citep{guo2026editr1}. At the system level, Swift-Image combines parallel expert RL with multi-teacher on-policy distillation to mitigate interference among heterogeneous objectives~\citep{hu2026swiftimage}. Together, these approaches motivate rewards that reliably distinguish successful edits from subtle preservation failures while remaining efficient enough for repeated online evaluation.

\paragraph{Regression-based Reward Modeling.}
Regression-based reward models learn holistic or dimension-specific preferences from vision-language representations, commonly through regression or pairwise preference objectives. ImageReward, PickScore, and HPS/HPSv2 model human preferences for text-to-image generation~\citep{xu2023imagereward,kirstain2023pickscore,wu2023hps,wu2023hpsv2}; MPS extends evaluation to multiple dimensions~\citep{zhang2024mps}, while HPSv3 explores broader visual preference modeling~\citep{ma2025hpsv3}. For image editing, HIVE and EditReward learn editing-specific preference signals~\citep{zhang2024hive,wu2026editreward}, and ADIEE develops automated editing evaluation~\citep{chen2025adiee}. Avoiding lengthy rationale generation makes direct scoring attractive for online RL. However, scores alone do not explicitly expose which concrete editing or preservation requirements are satisfied, motivating richer supervision during reward learning.

\paragraph{Generative Reward Modeling.}
Generative reward models make evaluation more explicit by producing rationales or structured critiques. EditScore, RationalRewards, SpatialReward, and UnifiedReward-Think investigate reasoning-based visual assessment from complementary perspectives~\citep{luo2026editscore,wang2026rationalrewards,long2026spatialreward,wang2025unifiedrewardthink}. Such reasoning can provide detailed diagnostic evidence, but autoregressive evaluation introduces additional computation when repeated across online RL samples. JRM and Z-Reward address this efficiency challenge by internalizing reasoning into more direct reward prediction pathways~\citep{yang2026jrm,jin2026zreward}. Our focus is on the supervision transferred into such efficient predictors: instance-specific verification requirements grounded in both the source image and the editing instruction.

\paragraph{Instance-Adaptive Verification.}
Adaptive evaluation tailors its criteria to individual inputs. RubricRM generates input-specific dimensions, weights, and scoring criteria~\citep{kan2026rubricrm}, while Edit-R1 decomposes editing instructions into principles for explicit verification~\citep{guo2026editr1}. Generated checklists have also been studied for language-model evaluation~\citep{cook2024tick}; in multimodal preference optimization, rDPO uses instance-specific checklist-style rubrics to construct and filter preference data~\citep{yu2026rdpo}. CheckDistill combines adaptive verification with efficient reward inference: source-image-grounded atomic checklists and their verification evidence serve as privileged information during reward-model training. Distilling this supervision into direct multidimensional score distributions enables reward prediction without checklist inputs or explicit reasoning at deployment.

\section{Checklist Construction and Judge Templates}
\label{app:en:annotation}

\subsection{Checklist Construction Template}

Given one or more reference images and an editing instruction, the checklist constructor generates atomic verification items for Instruction Following (IF), Attribute Preservation (AP), and Context Preservation (CP). Each item contains five key fields: \texttt{type}, indicating the evaluation dimension; \texttt{object}, specifying the edited or preserved entity; \texttt{location}, describing its spatial position; \texttt{bbox\_2d}, giving the normalized bounding box in the reference image; and \texttt{content}, specifying the atomic requirement to be verified. Bounding-box coordinates are normalized to integer values in $[0,1000]$, and \texttt{bbox\_2d} is set to \texttt{null} for newly added objects when no expected generation region can be determined. For multiple reference images, \texttt{ref\_image} further identifies the source image containing the object. The checklist is returned as a single JSON list, as specified in the template below.

\begin{promptbox}{Checklist Construction Template}
You are a strict and meticulous expert in designing image-editing checklists. Given one or more reference images and an editing instruction, generate checklists that can guide subsequent evaluation.

**## Input**
Prompt: {request}
Image input: one or more reference images <image>

**## Generation Procedure**
1. Inspect the reference image(s): describe key elements that are actually visible in the reference image(s), including their location, count, color, shape, pose, and other observable attributes. Do not infer or imagine any invisible content.
2. Parse the instruction: decompose the editing request into concrete add/remove/modify operations.
3. Generate checklists: decompose each operation into independently verifiable checklist items.

**## Checklist Types**
- Instruction Following (IF): the specific editing action that must be performed on an object involved in the edit.
- Attribute Preservation (AP): the key attributes of an edited object that should remain unchanged during editing.
- Context Preservation (CP): for each object not involved in the edit, set the content to "Keep this object unchanged."

**## Generation Rules (Must Be Followed)**
1. **Groundedness**: The object, location, and content in every checklist item must be directly grounded in either the visible content of the reference image(s) or the explicit editing instruction. 
Do not introduce objects, attributes, or locations that are unsupported by either the reference image(s) or the editing instruction.

2. **Atomicity**: Each checklist item must describe only one independently verifiable target. Do not combine multiple operations, objects, or attributes into a single item.

3. **Removal Operations**: Specify only the object to be removed and its location. Do not describe how the background should be filled after removal.

4. **BBox Localization**: Each checklist item must include `bbox_2d`, indicating the bounding box of the object in the reference image in the format [x1, y1, x2, y2], where the coordinates correspond to the top-left and bottom-right corners and are normalized to integer values in [0, 1000]. The `bbox_2d` must tightly cover the actually visible extent of the object. When multiple reference images are provided, additionally use `ref_image` to indicate the 1-based index of the reference image containing the object. For an IF item involving a newly added object that does not exist in the reference image, set `bbox_2d` to the expected generation region; if the expected region cannot be determined, set it to null.

5. **Self-Check**: Before output, verify that every checklist item is directly supported by either the reference image(s) or the editing instruction, and that each `bbox_2d` is consistent with the actual object location.

**## Output**
Write all checklist items into a single JSON list. The output language must be consistent with the editing instruction:
[
  {
    "checklist_id": IntType,
    "type": "IF | AP | CP",
    "object": "object name",
    "location": "position in reference image",
    "bbox_2d": [x1, y1, x2, y2],
    "content": "editing instruction or preservation constraint"
  },
  ...
]
\end{promptbox}

\subsection{Judge Template}
We use two complementary judges for semantic consistency (SC) and perceptual quality (PQ). The SC judge performs instance-specific, checklist-guided evaluation using the editing instruction, generated checklist, reference image(s), and edited result. It independently assesses Instruction Following (IF), Attribute Preservation (AP), and Context Preservation (CP), and outputs checklist-level judgments, dimension-wise scores and rationales, and an overall SC score. The PQ judge instead evaluates the edited result alone under a fixed quality rubric covering sharpness, generation artifacts, structural plausibility, lighting and perspective consistency, and boundary blending, and outputs a perceptual-quality score with a concise rationale. The complete prompt templates are provided below.

\begin{promptbox}{SC Judge Template}
You are a strict and consistent expert in image-editing quality evaluation. Your task is to assign absolute quality scores for instruction following and context preservation of an edited result.

**## Input**
Prompt: {request}
Checklists: {checklists_json}
Image input order: one or more reference images (from the first image to the second-to-last image) -> edited result image (the last image)

**## Evaluation Rules (Must Be Followed)**
1. **Checklist Evaluation**: Evaluate every provided checklist item one by one. Do not introduce additional evaluation criteria, and do not speculate about details that are not visible in the images.
2. **Visible Evidence Only**: Every judgment must be based only on differences that can be directly observed between the reference image(s) and the edited result. You may use the checklist `bbox_2d` ([x1, y1, x2, y2], normalized to [0, 1000]) to localize the corresponding object in the reference image before comparing it with the relevant region in the edited result. 
3. **Independent Scoring**: Score Instruction Following (IF), Attribute Preservation (AP), and Context Preservation (CP) independently. Performance in one dimension must not compensate for errors in another.
4. **Stronger Penalty for Incorrect Edits**: Editing the wrong object, introducing malformed structures, causing structural collapse, or violating physical plausibility should be penalized more heavily than simply failing to perform an intended edit.
5. **Full Satisfaction Criterion**: A checklist item is considered fully satisfied only when both conditions hold: the requested edit or preservation constraint is correctly satisfied, and the corresponding region is visually free of defects.

**## Scoring Scale (Shared by IF, AP, and CP; Integer Scores from 0 to 9)**
First estimate the fraction of checklist items satisfied within the dimension, then determine the score by considering the severity of the observed defects:
- 9: All requirements are fully satisfied, with no visible defects.
- 8: All requirements are satisfied, with only minor imperfections in a small number of non-critical aspects.
- 7: All critical requirements are completed, while a few non-critical aspects are not fully satisfied.
- 6: Most requirements are satisfied, with noticeable but non-severe defects.
- 5: Roughly two-thirds of the requirements are satisfied, or there is one clear defect.
- 4: Roughly half of the requirements are satisfied, or there are multiple clear defects.
- 3: Only a small portion is satisfied, or there is one severe defect, such as editing the wrong object, deformation, or structural collapse.
- 2: Very few requirements are satisfied, or there are multiple severe defects.
- 1: Almost none of the requirements are satisfied, with severe incorrect edits.
- 0: The requested edit or preservation constraint is entirely unfulfilled or completely incorrect.

**Discriminability Requirement**: Use the full 0-9 score range. Do not cluster scores around 5-7 without sufficient evidence. Scores must clearly distinguish among perfect results (9), minor imperfections (7-8), noticeable defects (4-6), and severe failures (0-3).

Final `score`: compute an overall score with IF as the primary factor and AP/CP as secondary factors. The score may be a floating-point value with one decimal place. If IF <= 3, the final score must not exceed 4.

**## Output (Strict JSON Only; Use the Same Language as the Editing Instruction)**
- `ckl reason`: provide a judgment rationale for every checklist item without omission.
- `IF reason`, `AP reason`, `CP reason`, and the overall `reason`: summarize all relevant observations within the corresponding dimension, using the object as the grammatical subject whenever possible.
- Order observations by their impact on the score, from largest to smallest. Keep the explanations concise and non-redundant.

{
  "ckl reason": "string",
  "IF score": <0-9 int>,
  "IF reason": "string",
  "AP score": <0-9 int>,
  "AP reason": "string",
  "CP score": <0-9 int>,
  "CP reason": "string",
  "score": <0-9 float>,
  "reason": "string"
}
\end{promptbox}
\begin{promptbox}{PQ Judge Template}
You are a strict and consistent expert in image quality evaluation. Your task is to assign an absolute score based solely on the visual quality of the edited result itself, independent of whether the editing instruction has been successfully fulfilled.

**## Input**
Prompt: {request}
Image input: edited result image

**## Evaluation Rules (Must Be Followed)**
1. Base all judgments only on directly observable evidence in the image. Do not speculate about nonexistent content, and do not over-penalize negligible imperfections.
2. Inspect the image in the following order:
   (*@\ding{172}@*) sharpness / noise / blur
   (*@\ding{173}@*) artifacts / compositing misalignment / unnatural textures
   (*@\ding{174}@*) deformation or structural errors in humans and objects
   (*@\ding{175}@*) consistency of lighting / perspective / color
   (*@\ding{176}@*) edge blending and overall realism
3. First identify the defects that are actually present, then assign a score according to the scale below.

**## Scoring Scale (Integer Scores from 0 to 9)**
- 9: Highly realistic and natural, with sharp image quality, no visible defects, and rich details.
- 8: Excellent quality, with only one minor imperfection that does not affect the overall viewing experience.
- 7: Good quality, with a small number of minor imperfections that do not materially affect the overall appearance.
- 6: Acceptable quality, with noticeable imperfections that slightly affect the viewing experience.
- 5: Average quality, with multiple imperfections or one clear defect.
- 4: Poor quality, with multiple clear defects, such as local artifacts or mild deformation.
- 3: Bad quality, with severe defects such as obvious deformation, structural errors, or large-scale artifacts.
- 2: Very poor quality, with multiple severe defects that substantially degrade the visual result.
- 1: Severely distorted, with major structural collapse and little practical usability.
- 0: Completely unusable, corrupted, or visually unrecognizable.

**Discriminability Requirement**: Use the full 0-9 score range. Do not cluster scores around 5-7 without sufficient evidence. Assign high scores (8-9) when no meaningful defects are present, and confidently assign low scores (0-3) when severe defects are observed.

**## Output (Strict JSON Only; Use the Same Language as the Editing Instruction)**
The `reason` field should briefly summarize the main defects or strengths that determine the score:
{
  "score": <0-9 int>,
  "reason": "string"
}
"""
\end{promptbox}

\section{Training Prompt Templates}
\label{app:en:prompts}

\subsection{Scoring Prompt for SFT and Student}

The same scoring prompt is used for Stage-I supervised fine-tuning and for the student model in Stage-II CR-OPSD. Given the editing instruction, one or more reference images, and the edited result, the model independently predicts four absolute scores for Instruction Following (IF), Attribute Preservation (AP), Context Preservation (CP), and Perceptual Quality (PQ). The model outputs the four integer scores in a fixed order, as specified below.

\begin{promptbox}{System Prompt}
You are a strict and consistent image editing quality judge. Based on the editing instruction (Prompt), compare the reference image(s) with the result image, and assign absolute quality scores along four independent dimensions: Instruction Following (IF), Attribute Preservation (AP), Context Preservation (CP), and Perceptual Quality (PQ).

**## Input**
Prompt: {request}
Image input order: one or more reference images (from the first to the second-to-last) (*@$\rightarrow$@*) the result image (the last one)
{image_tags}

**## Judging Rules (must be followed)**
1. **Visible facts only**: All scoring must be based on visual differences directly observable between the reference image(s) and the result image. Never speculate about details that do not exist in the images, and do not exaggerate minor flaws.
2. **Independent scoring**: IF, AP, CP, and PQ must each be scored independently, with no compensation between them.
3. **PQ is Prompt-independent**: PQ evaluates only the result image's own visual quality, independent of how well the editing instruction (Prompt) is fulfilled.
4. **Maintain score discrimination**: Make full use of the 0-9 score range and avoid clustering scores at 5-7 without justification. 9 indicates perfect performance, 7-8 indicates minor issues, 4-6 indicates obvious defects, and 0-3 indicates severe failure.

**## Evaluation Dimensions**
- **Instruction Following (IF)**: Evaluates whether the editing result fulfills the Prompt requirements. Focus on whether the edited object is correct, whether the edited content meets the requirements, whether the target attribute modifications are accurate, whether the degree of editing is sufficient, and whether there are missed edits, wrong edits, or clearly unreasonable results.
- **Attribute Preservation (AP)**: Evaluates whether the attributes of the edited object that the Prompt does not require to change remain consistent. Focus on whether non-target attributes such as object identity, appearance, color, structure, position, pose, quantity, proportion, material, and texture are preserved.
- **Context Preservation (CP)**: Evaluates whether objects not involved in the Prompt, the background, and the overall image properties such as color tone, lighting, and composition remain consistent with the reference image(s).
- **Perceptual Quality (PQ)**: Evaluates the result image's own visual quality and overall impression. Focus on whether the fundamental image quality (sharpness, blurriness, noise) is good; whether there are generation artifacts (unnatural textures, stitching misalignment, etc.), deformations, or physics-violating phenomena; whether human or object structures are plausible; whether lighting, shadows, perspective, reflections, and colors are harmonious and consistent; whether the background blends naturally with the subject; whether textures and edges look natural; and the overall realism, aesthetics, and detail quality.

**## Scoring Scale (integer, 0-9)**

**### Instruction Following (IF)**
9: Fully follows the Prompt. The edited object is correct, all requirements are accurately completed, and the result looks natural.
0: The edit is not performed at all, or is performed completely incorrectly: the edited object is wrong, the edited content is contrary or irrelevant to the Prompt requirements, or the editing result is severely deformed or structurally broken.

**### Attribute Preservation (AP)**
9: The edited object remains highly consistent; only the attributes explicitly required by the Prompt are changed, and all non-target attributes are preserved.
0: The non-target attributes of the edited object are severely lost; its identity or structure is completely changed and it can no longer be recognized as the same object in the reference image(s).

**### Context Preservation (CP)**
9: The non-edited content remains completely consistent, with no observable changes.
0: The non-edited content is largely destroyed or entirely repainted, with almost no consistency with the reference image(s).

**### Perceptual Quality (PQ)**
9: Realistic and natural; sharp fundamental quality with no blurring, noise, artifacts, deformations, or physics violations; plausible structure; harmonious lighting and color; rich detail.
0: Severely distorted; pervasive deformations, generation artifacts, quality defects, and physics errors; broken structure; essentially unusable.

**## Output:**
[IF score, AP score, CP score, PQ score]
\end{promptbox}

\subsection{Privileged Information for the Teacher}

In Stage-II CR-OPSD, the teacher receives the same evaluation instruction and image inputs as the student, together with additional instance-specific privileged information $z=(C,J,A)$. Specifically, $C$ denotes the adaptive checklist constructed for the current editing instance, $J$ contains checklist-level judgments grounded in visible evidence, and $A$ provides dimension-wise rationales for IF, AP, CP, and PQ. The additional teacher context is formatted as follows:

\begingroup\small
\begin{promptbox}{Privileged Information}

## Checklist

{For each checklist item: checklist_id, type, object, location, bbox_2d, content;
include ref_image when multiple reference images are provided.}

## Checklist Judgments

{For each checklist item, provide its judgment and corresponding visible evidence,
aligned by checklist_id.}

## Dimension-wise Rationales

IF: {Rationale for instruction following}
AP: {Rationale for attribute preservation of the edited object}
CP: {Rationale for preservation of non-edited content}
PQ: {Rationale for the perceptual quality of the edited result}

\end{promptbox}
\endgroup

\section{Additional Quantitative Results}

\subsection{Complete Ablation Results}
We provide the complete ablation results on MMRB2, MER-Bench, and EditReward-Bench. Table~\ref{tab:evidence} reports the effect of different privileged-information configurations, while Table~\ref{tab:reliability} presents the full ablation of design choices in CR-OPSD.

\begin{table}[!ht]
\centering
\caption{Privileged-information ablation (\%, $\uparrow$). Bold marks column maxima.}
\label{tab:evidence}
\setlength{\tabcolsep}{3pt}
\resizebox{\linewidth}{!}{%
\begin{tabular}{
l
r r >{\columncolor{gray!15}}r
r r r >{\columncolor{gray!15}}r
r r >{\columncolor{gray!15}}r
}
\toprule
& \multicolumn{3}{c}{\textbf{MMRB2}}
& \multicolumn{4}{c}{\textbf{MER-Bench}}
& \multicolumn{3}{c}{\textbf{EditReward-Bench}} \\
\cmidrule(lr){2-4}\cmidrule(lr){5-8}\cmidrule(lr){9-11}

\textbf{Privileged Information}
& \textbf{Single} & \textbf{Multi} & \textbf{Overall}
& \textbf{2 pair} & \textbf{3 pair} & \textbf{4 pair} & \textbf{Overall}
& \textbf{PF} & \textbf{CS} & \textbf{Overall} \\
\midrule

Checklist $C$
&72.99&\textbf{67.98}&72.10
&81.00&56.50&27.00&54.83
&\textbf{86.60}&84.20&82.10\\

+ Checklist Judgment $(C,J)$
&73.24&66.29&72.00
&81.50&59.50&27.00&56.00
&84.60&\textbf{85.10}&82.20\\

\rowcolor{methodblue}
+ Dimension-wise Rationale $(C,J,A)$
&\textbf{73.36}&67.42&\textbf{72.30}
&\textbf{82.00}&\textbf{60.50}&27.50&\textbf{56.67}
&85.50&84.10&\textbf{82.90}\\

+ Dimension Scores $(C,J,A,\mathbf g)$
&71.17&65.17&70.10
&\textbf{82.00}&57.50&\textbf{28.00}&55.83
&84.90&83.90&81.30\\

\bottomrule
\end{tabular}}
\end{table}

\begin{table}[!ht]
\centering
\caption{
Ablation study of the design choices in CR-OPSD (\%, $\uparrow$).
All variants use the same privileged information $(C,J,A)$.
Bold indicates the best result in each column.
}
\label{tab:reliability}
\setlength{\tabcolsep}{3pt}
\resizebox{\linewidth}{!}{%
\begin{tabular}{
l
r r >{\columncolor{gray!15}}r
r r r >{\columncolor{gray!15}}r
r r >{\columncolor{gray!15}}r
}
\toprule
& \multicolumn{3}{c}{\textbf{MMRB2}}
& \multicolumn{4}{c}{\textbf{MER-Bench}}
& \multicolumn{3}{c}{\textbf{EditReward-Bench}} \\
\cmidrule(lr){2-4}\cmidrule(lr){5-8}\cmidrule(lr){9-11}

\textbf{Variant}
& \textbf{Single} & \textbf{Multi} & \textbf{Overall}
& \textbf{2 pair} & \textbf{3 pair} & \textbf{4 pair} & \textbf{Overall}
& \textbf{PF} & \textbf{CS} & \textbf{Overall} \\
\midrule

w/o Reliability Weighting
&71.41&63.48&70.00
&81.50&57.50&26.00&55.00
&84.70&\textbf{85.10}&82.10\\

Linear Reliability Mapping
&72.51&65.73&71.30
&81.50&59.50&\textbf{27.50}&56.17
&85.00&84.70&82.30\\

Argmax-based Reliability
&71.90&64.61&70.60
&\textbf{82.50}&57.50&26.50&55.50
&84.60&84.20&82.60\\

\rowcolor{methodblue}
\textbf{CR-OPSD (Ours)}
&\textbf{73.36}&\textbf{67.42}&\textbf{72.30}
&82.00&\textbf{60.50}&\textbf{27.50}&\textbf{56.67}
&\textbf{85.50}&84.10&\textbf{82.90}\\

\bottomrule
\end{tabular}}
\end{table}

\subsection{Complete Image Editing Results}

We provide the complete results of the online RL experiments in Tables~\ref{tab:imgedit_bench} and~\ref{tab:redbench}.

\begin{table}[!ht]
\centering
\caption{Results on ImgEdit-Bench and GEdit-Bench-EN.}
\label{tab:imgedit_bench}
\setlength{\tabcolsep}{3pt}
\resizebox{\linewidth}{!}{%
\begin{tabular}{
l
rrrrrrrrr
>{\columncolor{gray!15}}r
rr
>{\columncolor{gray!15}}r
}
\toprule
& \multicolumn{10}{c}{\textbf{ImgEdit-Bench}}
& \multicolumn{3}{c}{\textbf{GEdit-Bench-EN}} \\
\cmidrule(lr){2-11}
\cmidrule(lr){12-14}

\textbf{Model}
& \textbf{Add}
& \textbf{Adjust}
& \textbf{Extract}
& \textbf{Replace}
& \textbf{Remove}
& \textbf{Background}
& \textbf{Style}
& \textbf{Compose}
& \textbf{Action}
& \textbf{Overall}
& \textbf{G\_SC}
& \textbf{G\_PQ}
& \textbf{G\_O} \\
\midrule

Qwen-Image-Edit
& 4.40 & 4.30 & 3.67 & 4.64 & 4.34
& 4.30 & 4.83 & 3.39 & \textbf{4.71} & 4.287
& 8.855 & 7.949 & 8.181 \\

UniWorld-Qwen-Image-Edit
& 4.42 & 4.31 & 3.71 & \textbf{4.69} & 4.32
& 4.28 & 4.83 & 3.42 & 4.68 & 4.296
& 8.824 & 7.884 & 8.119 \\

\rowcolor{methodblue}
\textbf{+ DiffusionNFT (CheckDistill-9B)}
& \textbf{4.47}
& \textbf{4.68}
& \textbf{4.52}
& 4.65
& \textbf{4.41}
& \textbf{4.40}
& \textbf{4.89}
& \textbf{3.60}
& 4.51
& \textbf{4.459}
& \textbf{8.909}
& \textbf{8.039}
& \textbf{8.280} \\

\bottomrule
\end{tabular}}
\end{table}

\begin{table}[!ht]
\centering
\caption{Results on RED-Bench.}
\label{tab:redbench}
\setlength{\tabcolsep}{2.5pt}
\resizebox{\linewidth}{!}{%
\begin{tabular}{
l
rrrrrrrrrrrrrrr
>{\columncolor{gray!15}}r
}
\toprule
& \multicolumn{16}{c}{\textbf{RED-Bench}} \\
\cmidrule(lr){2-17}

\textbf{Model}
& \textbf{Add}
& \textbf{Adjust}
& \textbf{Background}
& \textbf{Beauty}
& \textbf{Color}
& \textbf{Compose}
& \textbf{Extract}
& \textbf{Low-level}
& \textbf{Motion}
& \textbf{Portrait}
& \textbf{Remove}
& \textbf{Replace}
& \textbf{Stylize}
& \textbf{Text}
& \textbf{Viewpoint}
& \textbf{Overall} \\
\midrule

Qwen-Image-Edit
& 4.54
& 4.08
& 4.31
& 3.14
& 4.06
& 3.85
& 3.04
& 4.47
& \textbf{4.68}
& \textbf{4.37}
& 4.21
& \textbf{4.59}
& 4.68
& 3.94
& 3.55
& 4.10 \\

UniWorld-Qwen-Image-Edit
& 4.48
& 4.07
& 4.40
& 3.19
& 4.10
& 3.84
& 2.87
& 4.28
& 4.63
& 4.35
& 4.25
& 4.56
& 4.72
& 3.83
& 3.40
& 4.06 \\

\rowcolor{methodblue}
\textbf{+ DiffusionNFT (CheckDistill-9B)}
& \textbf{4.60}
& \textbf{4.25}
& \textbf{4.42}
& \textbf{3.37}
& \textbf{4.38}
& \textbf{3.99}
& \textbf{4.07}
& \textbf{4.59}
& 4.55
& 4.10
& \textbf{4.39}
& 4.56
& \textbf{4.73}
& \textbf{4.15}
& \textbf{3.73}
& \textbf{4.26} \\

\bottomrule
\end{tabular}}
\end{table}

\FloatBarrier
\begingroup
\setlength{\intextsep}{6pt plus 1pt minus 1pt}
\captionsetup[figure]{skip=4pt}
\section{Checklist-Guided Evaluation Examples}
We provide examples of CheckDistill's checklist-guided judgment in Figs.~\ref{fig:ckl_guide_1},~\ref{fig:ckl_guide_2}, and~\ref{fig:ckl_guide_3}.
Given an instance-specific checklist, CheckDistill performs fine-grained verification of each editing requirement and aggregates these item-level judgments into dimension-level assessments and scores.
\begin{figure}[H]
\centering
\includegraphics[width=\linewidth]{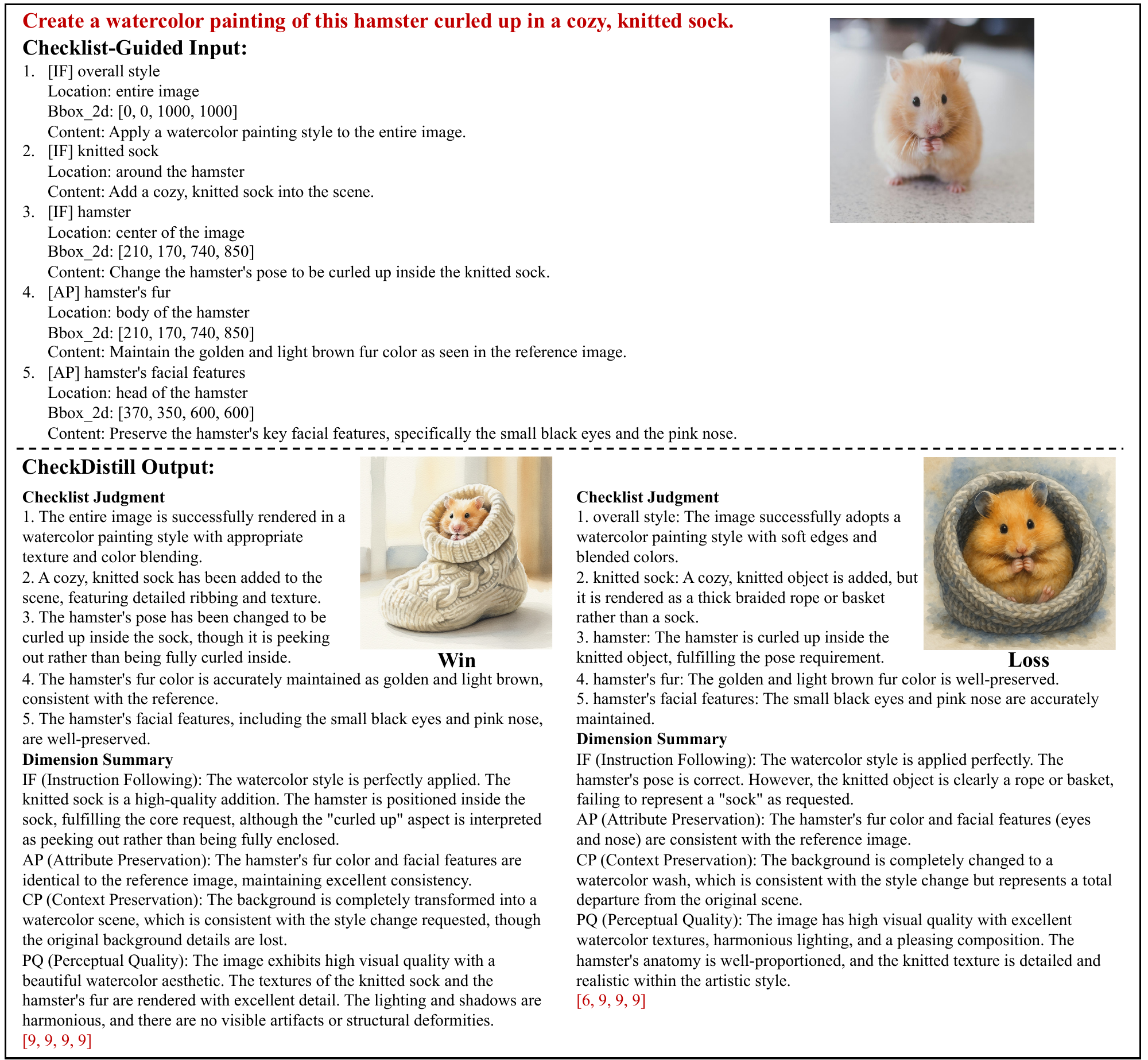}
\caption{Example of Checklist-Guided Judgment by CheckDistill}
\label{fig:ckl_guide_1}
\end{figure}

\begin{figure}[H]
\centering
\includegraphics[width=\linewidth]{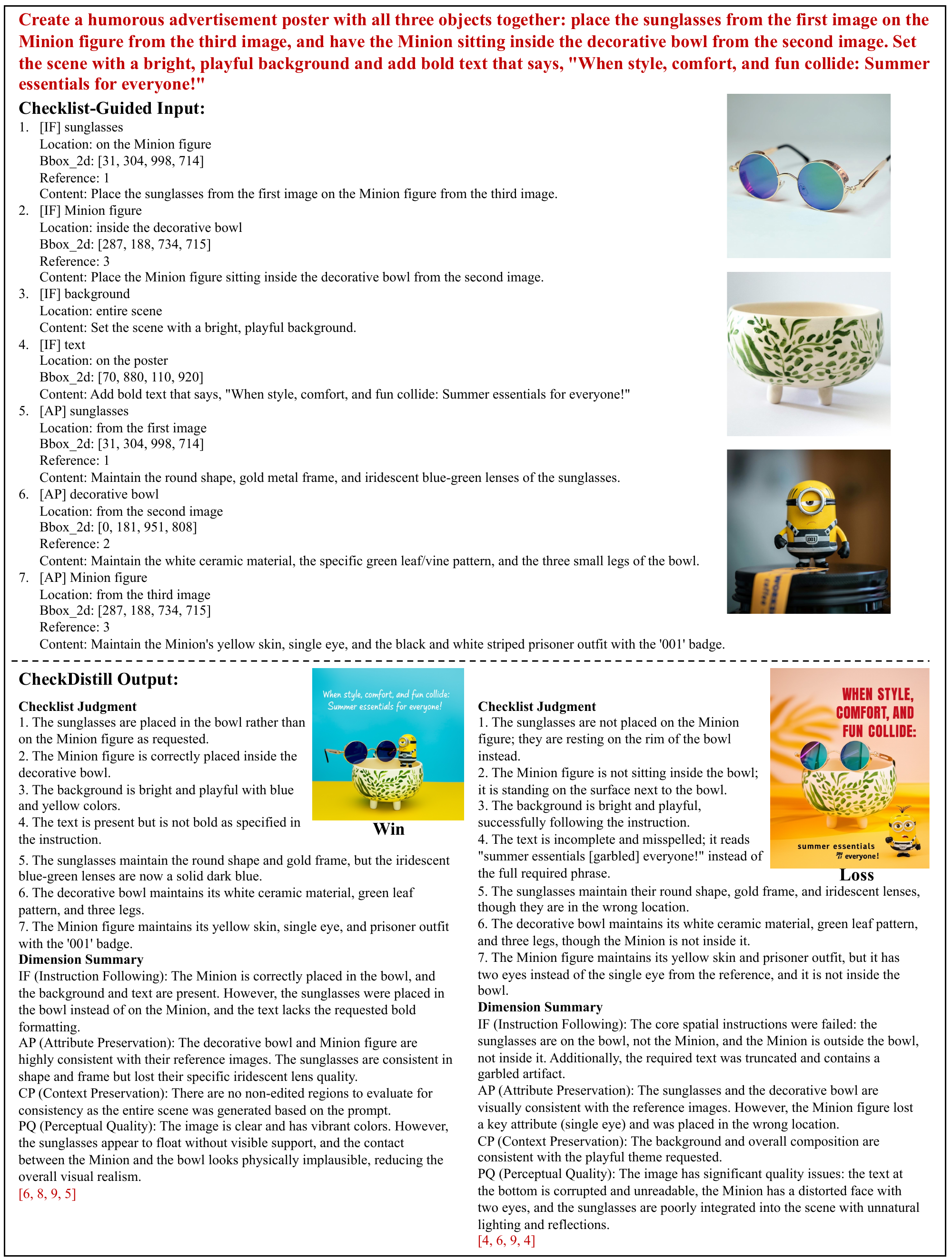}
\caption{Example of Checklist-Guided Judgment by CheckDistill}
\label{fig:ckl_guide_2}
\end{figure}

\begin{figure}[H]
\centering
\includegraphics[width=\linewidth]{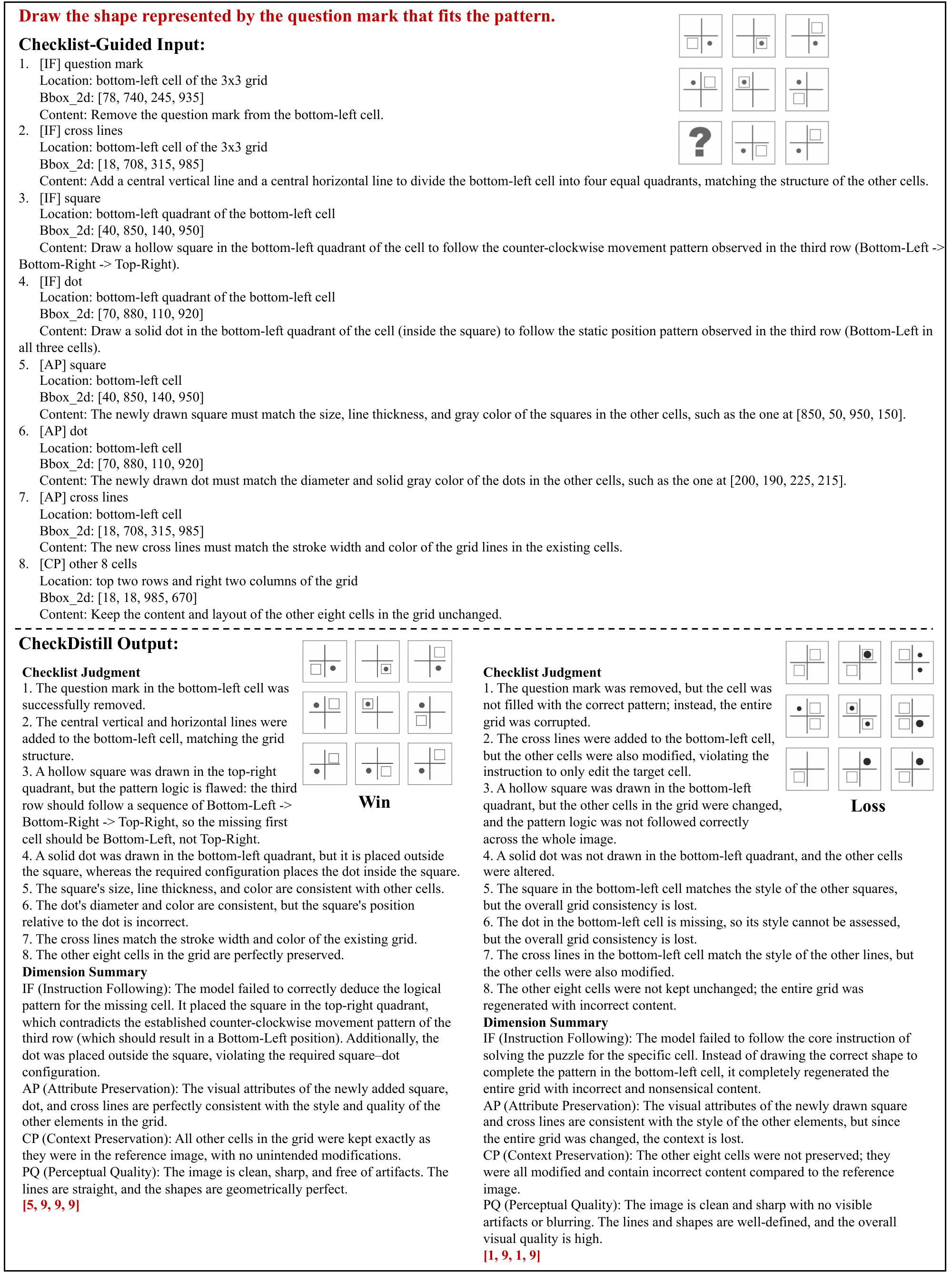}
\caption{Example of Checklist-Guided Judgment by CheckDistill}
\label{fig:ckl_guide_3}
\end{figure}

\Needspace{235pt}
\section{Additional Qualitative Evaluation}
We provide additional qualitative results to further illustrate the effectiveness of CheckDistill-guided RL. 
Figure~\ref{fig:reward_curve} presents the training reward curve, which steadily increases during optimization. Figure~\ref{fig:qualitative_1} and~\ref{fig:qualitative_2} provide qualitative comparisons before and after CheckDistill-guided RL, showing improved instruction following, stronger source-image consistency, and better perceptual quality.

\begin{figure}[H]
\centering
\includegraphics[width=0.44\linewidth]{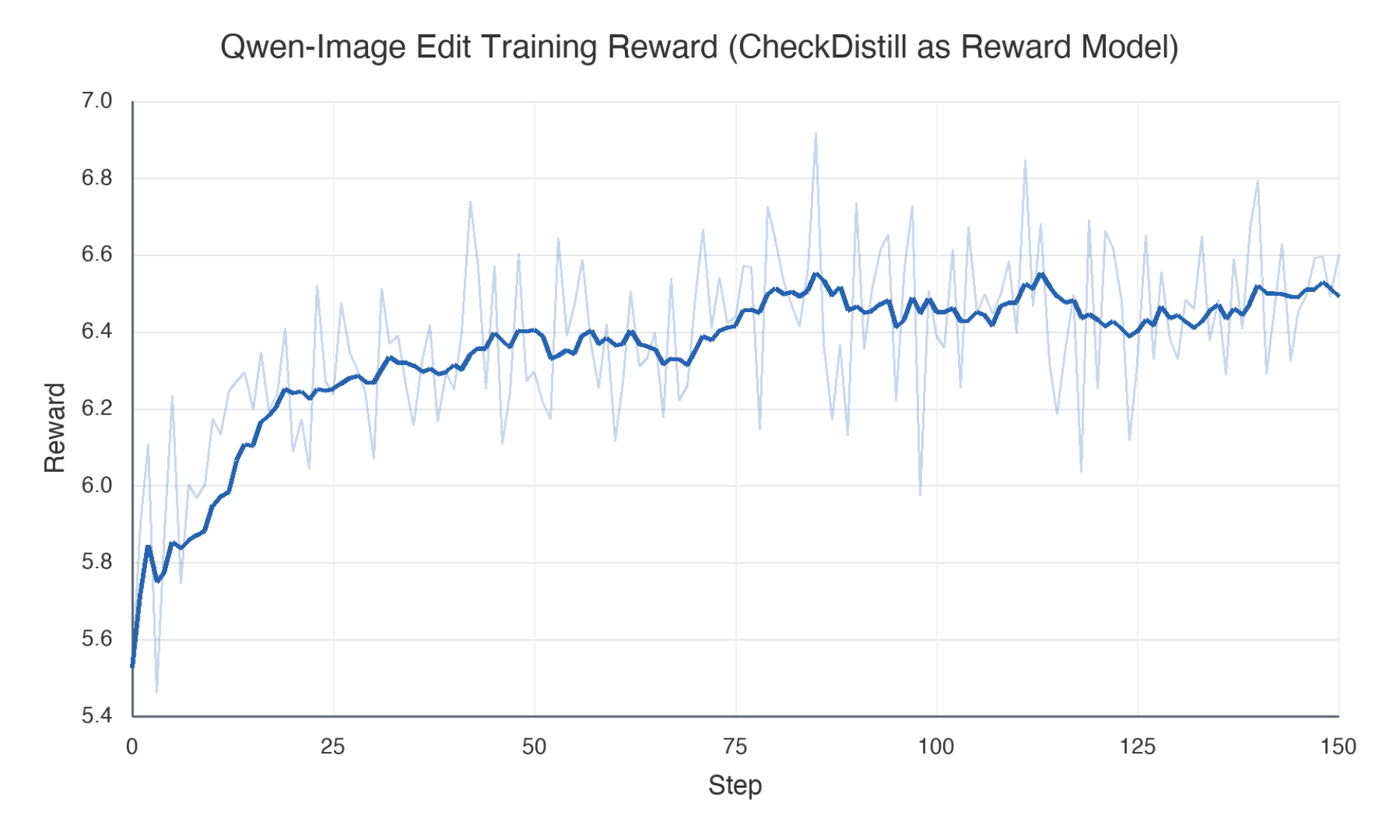}
\caption{Training reward for Qwen-Image-Edit using CheckDistill as the reward model.}
\label{fig:reward_curve}
\end{figure}

\begin{figure}[H]
\centering
\includegraphics[width=\linewidth]{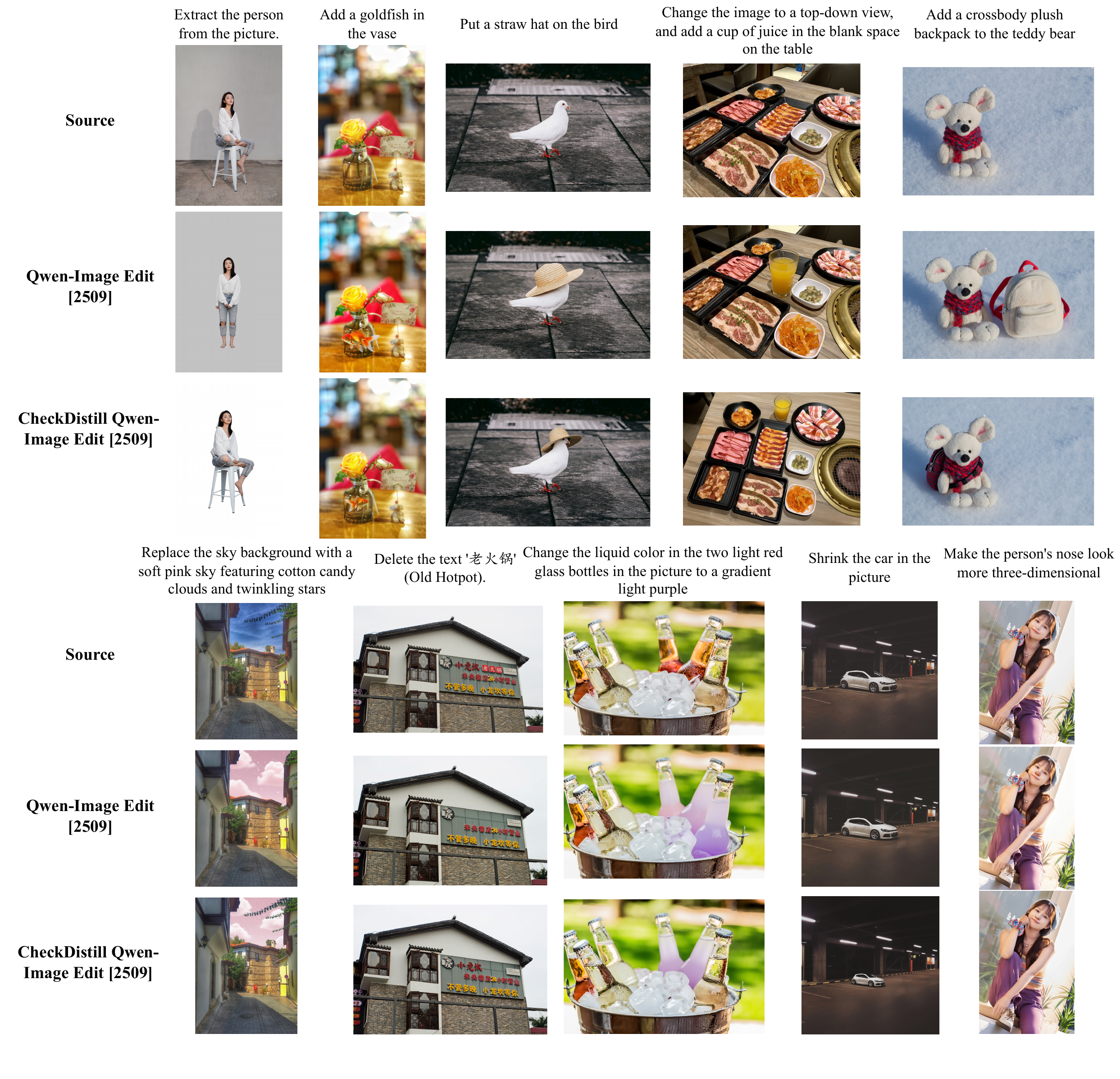}
\caption{Qualitative Comparison Before and After CheckDistill-based Reward Training.}
\label{fig:qualitative_1}
\end{figure}

\begin{figure}[H]
\centering
\includegraphics[width=0.78\linewidth]{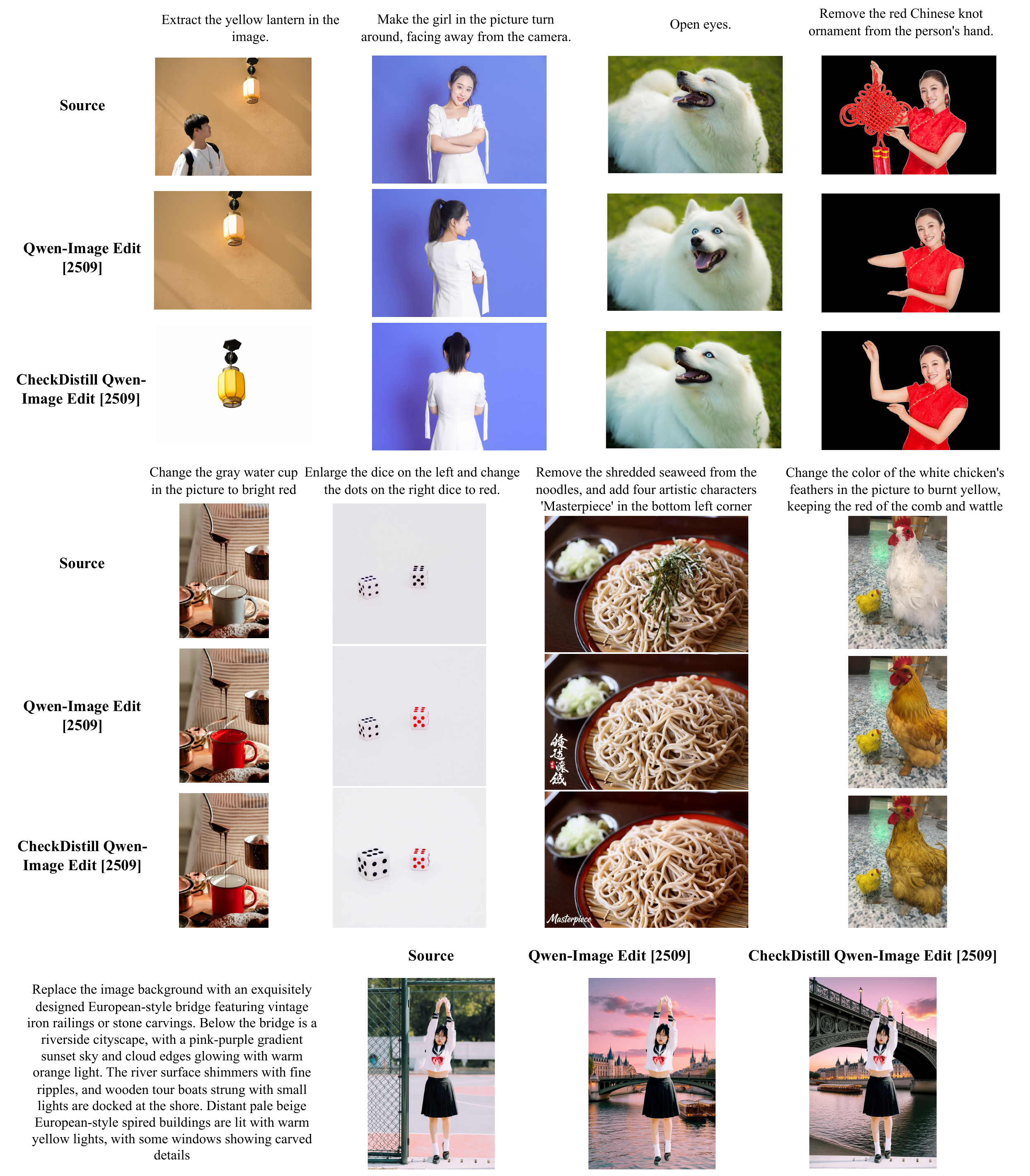}
\caption{Qualitative Comparison Before and After CheckDistill-based Reward Training.}
\label{fig:qualitative_2}
\end{figure}

\section{Discussion}
Z-Reward~\citep{jin2026zreward} and our CheckDistill both distill reasoning-supported assessment into direct scoring, but address different challenges.
Z-Reward focuses on representing preference uncertainty and fine-grained score differences through reasoning-conditioned score distributions.
\textbf{Our starting point is that image-editing evaluation requires instance-specific verification of both requested changes and preservation constraints.}
We therefore introduce Dynamic Checklist Scaling to translate each editing instruction and source image into atomic verification criteria, enriching the teacher context with checklists and their supporting judgements.

This focus on verification evidence also shapes our distillation design.
While Z-Reward transfers scoring knowledge from a large reasoning teacher to a compact student, we use on-policy self-distillation with an evidence-enriched teacher of the same architecture.
Distribution-Aligned Context Compression transfers the teacher's scoring behavior at student-generated prefixes, and reliability weighting modulates supervision at score positions according to teacher--label agreement, accounting for potentially incorrect teacher judgements.
\textbf{Our contribution centers on constructing and reliably internalizing instance-specific verification evidence}, enabling direct scoring without checklist inputs or explicit reasoning at inference time.
\endgroup

\end{document}